\documentclass{article}

\usepackage[preprint]{icml2026}
\icmltitlerunning{LightTAR: Light and Fast Target-Aware Representations for Multimodal Tabular Learning}

\usepackage{amsmath,amsfonts,bm}

\def\eqref#1{(\ref{#1})}
\def\1{\bm{1}}

\def\vp{{\bm{p}}}
\def\vq{{\bm{q}}}

\def\vt{{\bm{t}}}

\def\vw{{\bm{w}}}
\def\vx{{\bm{x}}}

\def\mP{{\bm{P}}}

\def\mR{{\bm{R}}}

\def\mV{{\bm{V}}}
\def\mW{{\bm{W}}}

\def\mY{{\bm{Y}}}
\def\mZ{{\bm{Z}}}

\DeclareMathAlphabet{\mathsfit}{\encodingdefault}{\sfdefault}{m}{sl}
\SetMathAlphabet{\mathsfit}{bold}{\encodingdefault}{\sfdefault}{bx}{n}

\newcommand{\R}{\mathbb{R}}

\DeclareMathOperator*{\argmax}{arg\,max}

\usepackage{amsmath}
\usepackage{amssymb}
\usepackage{amsfonts}
\usepackage{amsthm}
\usepackage{graphicx}
\usepackage{stfloats}
\usepackage{booktabs}
\usepackage{tikz}
\usetikzlibrary{positioning, arrows.meta}
\usepackage{hyperref}
\hypersetup{hidelinks}
\usepackage{url}
\usepackage{microtype}
\usepackage{xcolor}

\newcommand{\changed}[1]{#1}
\newcommand{\newtext}[1]{#1}
\newcommand{\changedon}{}
\newcommand{\unreviewedon}{}
\newcommand{\markoff}{}

\begin{document}

\twocolumn[
  \icmltitle{LightTAR: Light and Fast Target-Aware Representations\\
  for Multimodal Tabular Learning}

  \begin{icmlauthorlist}
    \icmlauthor{Eilam Shapira}{technion}
    \icmlauthor{Alan Arazi}{technion}
    \icmlauthor{Shoham Grunblat}{technion}
    \icmlauthor{Roi Reichart}{technion}
  \end{icmlauthorlist}
  \icmlaffiliation{technion}{Technion -- Israel Institute of Technology}
  \icmlcorrespondingauthor{Eilam Shapira}{eilam.shapira@gmail.com}
  \icmlkeywords{multimodal tabular learning, frozen representations, supervised dimensionality reduction}
  \vskip 0.3in
]
\printAffiliationsAndNotice{}

\begin{abstract}
\changed{Multimodal tabular learning involves classification and regression
on tables combining structured numerical and categorical features with
unstructured text and images. These unstructured fields are commonly
represented by pretrained encoder embeddings reduced with principal component
analysis (PCA). This reduction prioritizes input variance but may overlook
information relevant to predicting the target.}
\changed{We introduce LightTAR,
a simple, inexpensive method that adapts dimensionality reduction to the
prediction task. LightTAR augments the standard PCA representation with
a few target-aware features extracted jointly from the final and an
intermediate encoder layer using regularized partial least squares (PLS).}
\changed{We evaluate LightTAR with 21 tabular models, including tree ensembles,
neural networks, and tabular foundation models. LightTAR improves average
predictive performance for all models and, averaged across models, improves
prediction on almost every MulTaBench dataset. These gains come with a small
increase in total runtime relative to PCA representations.}
\changed{LightTAR integrates directly into existing encoding pipelines,
improving tabular learners' predictive performance without modifying
the encoder or the learner.\footnote{\raggedright
Code is available at
\url{https://github.com/eilamshapira/LightTAR}.}}
\end{abstract}

\section{Introduction}
\label{sec:intro}

\begin{figure*}[!t]
\centering
\includegraphics[width=\linewidth]{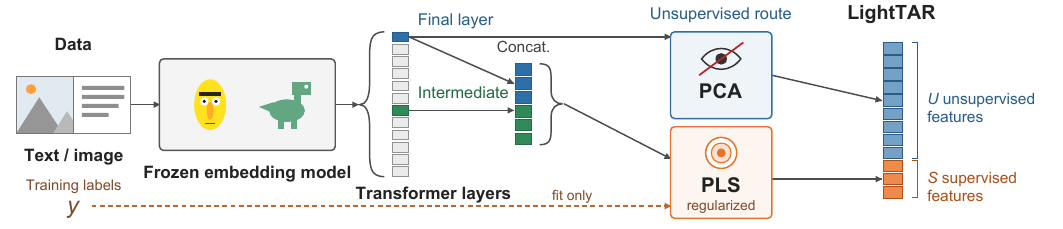}
\caption{\textbf{The Illustrated LightTAR Algorithm.} A frozen
embedding model encodes the text or image column. PCA reduces its
final-layer representation to $u$ features, while regularized PLS
uses training labels during fitting to extract $s$ target-aware features
from the concatenated final and intermediate layers.}
\label{fig:pipeline}
\end{figure*}

Tabular foundation models (TFMs) have recently established the
state of the art in tabular learning \citep{erickson_tabarena_2025},
learning generalizable representations of numerical and categorical
data through pretraining \citep{hollmann_accurate_2025,
qu_tabiclv2_2026, ma_tabdpt_2025}. Real tables, however, often carry
free text and images alongside their numerical and categorical columns.
For example, a dermatology record might combine a patient's age, a photo
of a lesion, and the clinician's notes, while a used-car listing might
pair its price and model year with photos of the car and the seller's
description. Predicting from such tables is known as
\emph{Multimodal Tabular Learning} (MMTL)
\citep{huang_fusion_2020, arazi_multabench_2026}.

While numerical and categorical features are consumed by a tabular
learner directly, text and images must first be encoded. MMTL pipelines
usually do this with a pretrained embedding model. This approach has
two main limitations. First, the resulting representations are
general-purpose and not adapted to the prediction task's labels. Since
pretrained encoders are trained to serve general-purpose objectives,
their representations encode diverse information, some of which might
be irrelevant to the task at hand. Second, these representations are
typically high-dimensional, making them computationally costly for
tabular learners. In particular, a pretrained encoder typically
represents text or images as a vector of several hundred dimensions,
so a single column can contribute more features than the rest of the
table holds.
A gradient-boosted ensemble evaluates candidate features at every split,
so its training cost grows with the number of columns. With current
tabular foundation models, the problem can be even more severe: they
struggle with large feature counts and, in some cases, cannot handle
them at all \citep{hollmann_accurate_2025}.

To address the high dimensionality of these representations, conventional
MMTL pipelines reduce the embeddings before the learner sees them
\citep{grinsztajn_vectorizing_2023, blayer2026strable}.
Conventionally, this reduction is unsupervised and, as such, is not
adapted to the prediction task. Each text or image value is encoded
using a pretrained encoder, and PCA is then fitted over the embeddings
of the column's training examples. Only a small number of components
explaining most of the input variance are retained. We call this
conventional approach the \emph{unsupervised pipeline}. Its features can be handed to a wide range
of tabular learners, including tree ensembles such as XGBoost
\citep{chen_xgboost_2016} and CatBoost
\citep{prokhorenkova_catboost_2018}, neural tabular architectures
such as RealMLP \citep{holzmuller_better_2024} and TabM
\citep{gorishniy_tabm_2025}, and tabular foundation models such as
TabPFN \citep{hollmann_accurate_2025}, TabICLv2
\citep{qu_tabiclv2_2026}, and Kumo-Tabular
\citep{qu_kumo_tabular_2026}.

Although PCA reduces the dimensionality of the representations, it
selects projection components based on input variance, without
considering their relationship to the target \citep{jolliffe1982note}.
As a result, information important for prediction may be discarded
during dimensionality reduction. Since usually only a few dozen
components are kept, a direction explaining little input variance
may be discarded even when it contains information highly correlated
with the task labels, such as a rare phrase in text or a small visual
distinction. Using the task labels to select the projection directions
instead yields a \emph{Target-Aware Representation} (TAR).
\citet{arazi_tabstar_2025, arazi_multabench_2026} obtain one by
fine-tuning the encoders against the task labels, improving the
performance of predictive models on a considerable share of the datasets
in common multimodal benchmarks. However, fine-tuning an encoder remains
slow and costly, and still risks overfitting, even when the model is
small and only a few layers are adapted using LoRA \citep{hu_lora_2021}.

Beyond how the dimensionality of the representations is reduced, the
encoder layers from which information is extracted also matter. In the
conventional pipeline, the representation of a text or image column
is typically taken from the encoder's final layer. This layer is trained
to place inputs that share a meaning close together, whether through
contrastive objectives over text pairs
\citep{reimers2019sentence, wang_text_2024} or through invariance to
augmented views of an image \citep{simeoni_dinov3_2025}. Such objectives
encourage the encoder to retain information relevant to its pretraining
objectives, but not necessarily the distinctions needed for the tabular
prediction task. As a result, the final layer may not retain task-relevant
information, such as exact wording in text
\citep{weller_theoretical_2025} or the presence and position of a small
visual detail in an image
\citep{pantazopoulos_lost_2024, tong_eyes_2024}. Previous studies have
shown that representations taken from intermediate encoder layers may
add information useful for prediction beyond what is available at the
final layer \citep{skean2025layer}, and that information no longer
accessible at the encoder's output may remain available in earlier
layers \citep{peters2018deep, tenney2019bert, hariharan2015hypercolumns}.
Using representations from multiple encoder layers could allow this
information to be exploited, but also increases the number of
dimensions that must subsequently be reduced.

Other approaches use large language models (LLMs) to construct
representations adapted to the prediction task
\citep{schindler2026case, lefebvre2026tile}. However, these approaches
require LLM inference to construct their representations, making feature
extraction more computationally demanding than with small embedding
encoders. We therefore seek a way to obtain target-aware representations
with minimal computational overhead beyond what is
required to encode the data using a pretrained encoder.

We introduce \textbf{LightTAR}: \textbf{Light} and Fast
\textbf{T}arget-\textbf{A}ware \textbf{R}epresentations. LightTAR
augments the PCA representation produced by the unsupervised pipeline
by appending a small number of features adapted to the prediction task.
These features are obtained by projecting onto directions selected
according to their covariance with the target, using Partial Least
Squares (PLS) with regularization that also favors retaining input
variance. The directions are extracted jointly from the encoder's
final layer and an intermediate layer, which may contain information
relevant to prediction that is not accessible at the final layer
(Figure~\ref{fig:pipeline}). The projection is constructed
deterministically, requires no gradient steps and no GPU, and takes
seconds.

We evaluate LightTAR on two complementary benchmarks. The core text
and image suites of MulTaBench \citep{arazi_multabench_2026}
comprise tasks where the additional modality carries predictive
information and adapting its representation to the task may improve
performance. In contrast, STRABLE \citep{blayer2026strable} includes
tables with string columns as they occur in the original data,
without selecting datasets based on the predictive contribution of
these columns or the benefits of adapting their representations to the
task. It therefore provides a control for evaluating LightTAR without
assuming that target-aware representations improve predictive
performance. These columns may contain free text, but also names,
identifiers, and structured codes, for which simple lexical
representations may remain competitive.

On MulTaBench, LightTAR improves average performance of all
21 tabular learners, from tree ensembles to the leading
tabular foundation models. Averaged across models, it improves
predictive performance on 19 of the 20 text datasets and 17 of the
20 image datasets. On STRABLE, where string columns were not selected
for their predictive contribution, the results are more modest and mixed.

LightTAR improves the representations used by tabular learners without
changing the encoder or the downstream learner. It adds little to the
cost of constructing the representations and can be integrated directly
into existing pipelines. These results highlight the potential of
adapting the dimensionality reduction step to the prediction task as
a simple, efficient way to improve multimodal tabular learning.

\unreviewedon
\section{Related Work}
\label{sec:related}
\changedon

We review the work LightTAR builds on: the models that learn from
multimodal tables, the methods that represent their text and image
columns, and the methods that reduce those representations.

\paragraph{Multimodal Tabular Learning Models.}
The field has not settled on how text and images should reach a tabular
predictor. One line of work serializes each row into a sentence and hands
it to a language model \citep{hegselmann_tabllm_2023,
schindler_tabgemma_2025, gardner_large_2024}. These approaches are particularly
effective in the few-shot regime, but scaling them to larger tables is
limited by the cost of running the language model and the context length
needed to represent the rows. They have yet to demonstrate a consistent
advantage on standard tabular benchmarks. Others fuse the modalities inside the predictor
\citep{spinaci_contexttab_2025, kim_multimodalpfn_2025, luo_time_2025},
though they build their representation without ever consulting the
target. A different line fine-tunes the encoder for each prediction task
\citep{arazi_tabstar_2025, tang_autogluon-multimodal_2024}. However,
fine-tuning is computationally expensive, and its gains are inconsistent
across datasets \citep{arazi_multabench_2026}.
CatBoost \citep{prokhorenkova_catboost_2018} includes built-in mechanisms
for creating target-aware text features and processing precomputed
dense representations. The dominant practice instead keeps
the modalities apart: each text or image value is encoded upstream and
appended to the table, so any tabular learner can consume it.
LightTAR strengthens this pipeline,
changing only how a text or image column is represented and leaving the
rest of the table and the predictor untouched. It thus applies to
the learners that currently lead TabArena \citep{erickson_tabarena_2025}.

\paragraph{Representing Text and Image Columns.}
The simplest representations of a string column are lexical, encoding
dirty or high-cardinality strings through character $n$-grams
\citep{cerda2018similarity}. Using $n$-grams produces a wide, sparse
feature matrix, so its dimensionality must be reduced before it is
passed to the tabular learner. These representations can suffice when
the strings are short, noisy, or near-duplicated
\citep{grinsztajn_vectorizing_2023, blayer2026strable}, but they are
limited in representing complex semantic relationships and cannot be
applied directly to images. Dense
representations instead encode each value with a pretrained model, such
as e5 for text \citep{wang_text_2024} or DINOv3 for images
\citep{simeoni_dinov3_2025}, producing high-dimensional, general-purpose
vectors that can serve diverse downstream applications, such as retrieval
and semantic similarity, and are not initially tailored to predicting
the target in the tabular task at hand. Although combining
representations from several layers has been shown to help in
non-tabular tasks \citep{hariharan2015hypercolumns, evci2022head2toe},
the conventional tabular pipeline typically uses only the final layer
to limit the representation's dimensionality.
To adapt the representation to the prediction task, target-aware
approaches use training labels at different stages: fine-tuning the
encoder, constructing the context in which the representation is
computed, or processing the representations after extraction.
\newtext{TILE \citep{lefebvre2026tile} and CASE
\citep{schindler2026case} use large language models and table-row
context to construct their representations. Computing representations
this way requires much more computation than using small embedding
encoders. TILE uses row context and intermediate-layer representations.
CASE continues pretraining an existing language model for tabular
prediction tasks, then produces row representations conditioned on
labeled examples from the training table.}
A cheaper approach leaves the encoder unchanged and instead processes
the representations it produces. \citet{bergstrom_comet_2026} learn to pool
token representations from the final layer of a frozen encoder into a
single vector. The pooling weights are learned from scores produced by
a tabular foundation model using labeled examples, without updating the
encoder's weights. After pooling, the vector's dimensionality is reduced
using unsupervised PCA.
LightTAR uses labels in the dimensionality reduction step itself,
combining information from the final and an intermediate layer of a
frozen encoder.

\paragraph{Supervised Dimensionality Reduction Methods.}
\label{sec:related_dimred}
Using labels to choose a low-dimensional representation has a long
history, including methods such as linear discriminant analysis (LDA)
\citep{fisher1936lda} and sliced inverse regression (SIR)
\citep{li1991sir}.
PCA is the conventional choice, ranking directions by input variance
whether or not they align with the target
\citep{grinsztajn_vectorizing_2023, blayer2026strable}. PLS instead
selects directions according to their covariance with the target
\citep{wold2001pls}.
\newtext{PLS also serves as a predictive model for high-dimensional
data, for example for prediction from spectra in RamanBench
\citep{koddenbrock2026ramanbench}.}
Supervised projections of pretrained features have been studied for
vision and text \citep{heras2020supervised, kumar2026supervised,
ressel2025linear, luo2025factor}.
LightTAR integrates supervised dimensionality reduction into a pipeline
for representing text and image columns: it augments PCA components
with target-aware directions based on regularized PLS. We
systematically evaluate this combination with a range of learners on
tabular tasks involving text or images.
\markoff

\markoff
\unreviewedon
\providecommand{\Lset}{\mathcal{L}}
\unreviewedon
\section{LightTAR}
\label{sec:method}

LightTAR augments the standard unsupervised representation with features
selected using the training labels, without changing the encoder or the
downstream learner. For each text or image column, it combines a PCA
\emph{anchor} with a small set of target-aware directions
(see Figure~\ref{fig:pipeline}). These directions are fitted by inexpensive
linear algebra on already-computed embeddings. We construct the
feature space, select the directions, and finally combine the two
blocks. The complete algorithm is given in Appendix~\ref{app:algo}.

\subsection{Constructing the Feature Space}
\label{sec:method_layers}

LightTAR uses multiple frozen encoder layers, since intermediate
representations may retain information less accessible at the output
\citep{skean2025layer}. Let $\phi^{(\ell)}(v)\in\R^{d_\ell}$
denote the encoder representation of a column value $v$ at layer
$\ell$. Given selected layers
$\Lset=\{\ell_1,\ldots,\ell_r\}$, we concatenate their vectors:
\begin{equation}
\label{eq:lighttar_main_layers}
\vx_i=\big[\phi^{(\ell)}(v_i)\big]_{\ell\in\Lset}
\in\R^D,\quad D=\sum_{\ell\in\Lset}d_\ell.
\end{equation}
For $n$ training examples, we standardize these vectors to obtain
the feature matrix $\tilde{\mZ}$. We select an \emph{anchor
layer} $\ell_a\in\Lset$ and fit its top $u$
PCA directions $\mV_u$ on the training data, where $u$ is the number
of unsupervised features.

\subsection{Extracting Target-Aware Directions}
\label{sec:method_supervised}

Given $\tilde{\mZ}$ and standardized training targets $\mY$
(class indicators for classification), let $s$ be the number of
supervised features. LightTAR extracts these $s$ directions
using a PLS objective to which we add an input-variance
term for regularization. We extract them sequentially,
removing the contribution of each direction from the features and
targets before selecting the next, so each can capture additional
signal.

A projection chosen solely from correlations with the training
labels can amplify chance associations, especially when labeled
data is scarce. Such associations may also lead the downstream
learner to overemphasize the encoded column rather than complementary
numerical or categorical features. We therefore temper the supervised
criterion with input variance to mitigate this risk, without
constraining how the learner combines its inputs.
At iteration $j$, let $\tilde{\mZ}^{(j)}$ and $\mY^{(j)}$
be the remaining feature and target matrices, respectively,
with $\tilde{\mZ}^{(1)}=\tilde{\mZ}$ and $\mY^{(1)}=\mY$.
Their feature--target and feature covariance matrices are
\[
\widehat S_{xy}^{(j)}
=\frac{1}{n}\tilde{\mZ}^{(j)\top}\mY^{(j)},\quad
\widehat S_{xx}^{(j)}
=\frac{1}{n}\tilde{\mZ}^{(j)\top}\tilde{\mZ}^{(j)}.
\]
We select
\begin{equation}
\label{eq:lighttar_main_objective}
\vw_j=\argmax_{\|\vw\|_2=1}
\underbrace{\|\widehat S_{xy}^{(j)\top}\vw\|_2^2}_{\text{PLS term}}
+\gamma\,\underbrace{\vw^\top\widehat S_{xx}^{(j)}\vw}_{\text{Regularization}} .
\end{equation}
The second term favors directions with high input variance,
independently of the labels. Its weight $\gamma\geq0$ controls
the trade-off: $\gamma=0$ recovers ordinary PLS, while larger
values give more weight to input variance.

The directions are learned from progressively deflated features.
After selecting $\vw_j$, we compute its scores
$\vt_j=\tilde{\mZ}^{(j)}\vw_j$ and the corresponding feature
loading $\vp_j=\tilde{\mZ}^{(j)\top}\vt_j/(\vt_j^\top\vt_j)$.
The product $\vt_j\vp_j^\top$ reconstructs the feature variation
captured by this direction. We subtract it (and update the targets
similarly) before selecting the next direction. Thus, each
$\vw_j$ is fitted to a different residual matrix.

To apply all directions to a new, non-deflated embedding, we
combine them into one projection. With
$\mW=[\vw_1,\ldots,\vw_s]$ and
$\mP=[\vp_1,\ldots,\vp_s]$, this projection is
$\mR=\mW(\mP^\top\mW)^{-1}$, which accounts for the
preceding deflation steps, so new examples require only a single
matrix multiplication.

\subsection{The Final Representation}
\label{sec:method_representation}

For a new column value $v$, let $\tilde{\vx}(v)$ denote its
standardized multi-layer vector and
$\tilde{\vx}^{(\ell_a)}(v)$ its standardized anchor-layer vector.
LightTAR combines the two blocks:
\begin{equation}
\label{eq:lighttar_main_representation}
\mathrm{L\text{-}TAR}(v)=
\big[
\mV_u^\top\tilde{\vx}^{(\ell_a)}(v)
\;\big\|\;
\mR^\top\tilde{\vx}(v)
\big]\in\R^{u+s}.
\end{equation}
For each text or image column, LightTAR produces $u+s$ features,
which are combined with the original numerical and categorical
features and passed to the downstream tabular learner. All
transformations are fitted on the training data and applied
unchanged to held-out examples.

\markoff

\markoff
\unreviewedon
\unreviewedon
\section{Experimental Setup}
\label{sec:setup}

\begin{figure*}[!t]
\centering
\IfFileExists{figures/elo_text_image.pdf}{%
\includegraphics[width=0.8\linewidth]{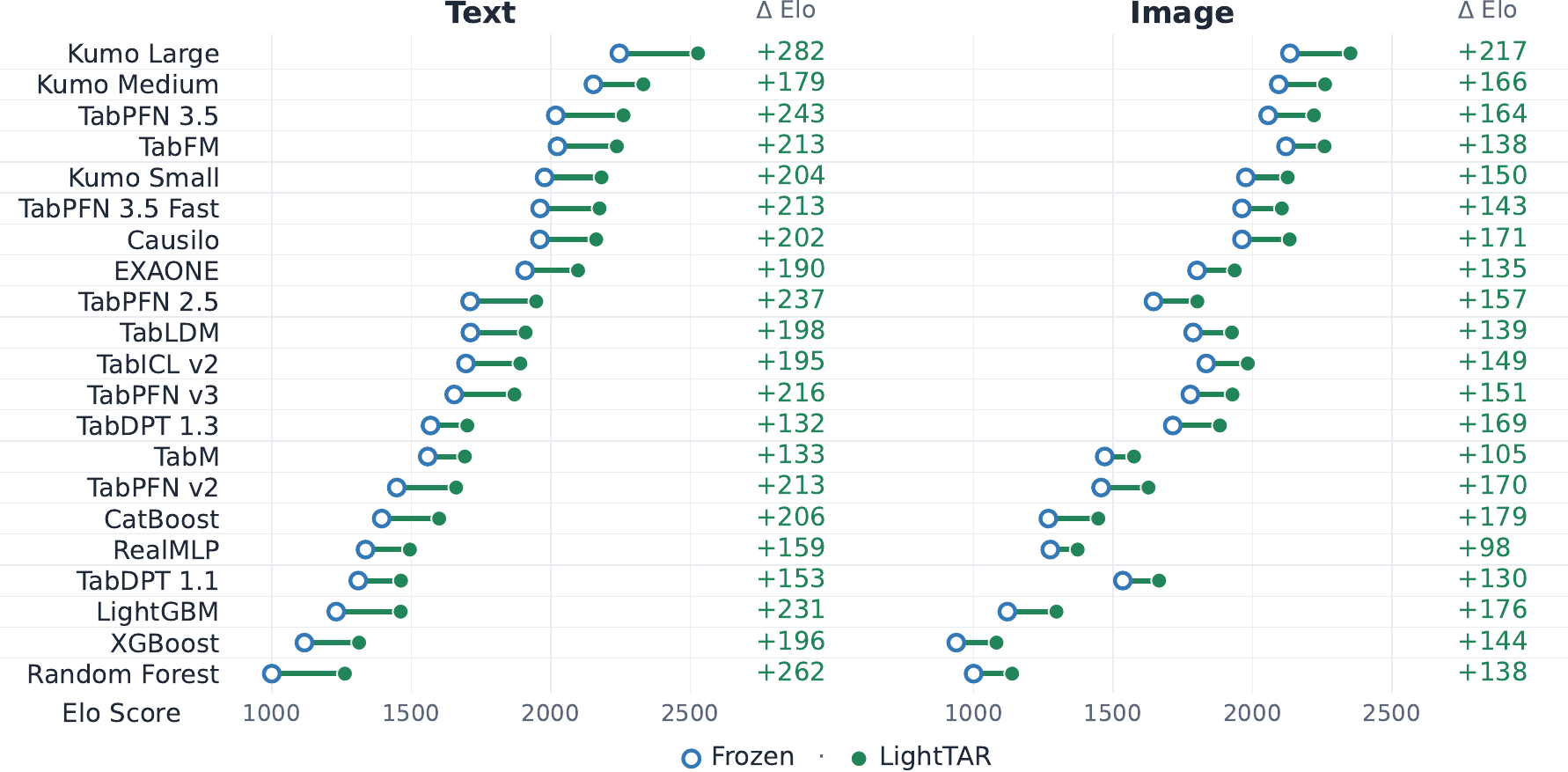}}{%
\framebox[\linewidth]{\rule{0pt}{20em}\texttt{paper/figures/elo\_text\_image.pdf} pending}}
\caption{\newtext{\textbf{Elo ratings of Frozen and
LightTAR on MulTaBench text and image.}
Elo ratings are computed separately for each
benchmark, with Random Forest Frozen anchored
at 1000.}}
\label{fig:panel}
\end{figure*}

\subsection{Benchmarks}
\newtext{We evaluate LightTAR on the 40 core MulTaBench datasets
\citep{arazi_multabench_2026}, comprising 20 text datasets and 20 image
datasets. The text suite includes 10
classification and 10 regression datasets, while the image
suite includes 12 classification and 8 regression datasets.
MulTaBench consists of tasks where the additional modality
carries predictive information and task-informed
adaptation can improve over unsupervised representations.}

\newtext{For a new prediction task, it is unclear in advance whether
adapting the representation using training labels will improve predictive
performance. We therefore also evaluate LightTAR on STRABLE
\citep{blayer2026strable}, whose 108 tables with string columns
(32 classification and 76 regression tasks) were selected without regard
to whether such adaptation improves prediction. STRABLE retains string
columns with minimal preprocessing. Many contain short, recurring values
such as names, dates, and structured codes.}

\subsection{Experimental Configuration}
\label{sec:setup_baselines}

\paragraph{Representations and encoders.}
\newtext{Following the MulTaBench paper,
we use \textit{e5-small-v2} for text and
\textit{DINOv3 ViT-S/16} for images. Our \textit{Frozen}
baseline reduces the final-layer embeddings to
30 PCA components per text or image column.
LightTAR retains this PCA anchor
($u=30$) and adds target-aware directions,
as described in Section~\ref{sec:method}.
Both methods keep the encoder weights fixed.}

\newtext{For LightTAR, we use the final and middle
hidden layers, $\Lset=\{12,6\}$, with the final
layer providing the PCA anchor. We set the number
of supervised directions to $s=5$ and the
regularization strength to $\gamma=60/n$, where
$n$ is the number of training examples used to
fit the projection. The layer selection, number of supervised
directions, and regularization strength were
chosen in preliminary experiments on a small
subset of text datasets and tabular models,
and kept fixed across datasets and modalities. In
Appendix~\ref{app:mwindow}, we examine how LightTAR's gain over Frozen
varies with $s$ and $\gamma$.}

\paragraph{Downstream models.}
\label{sec:setup_models}
\newtext{We evaluate 21 tabular models, including tree ensembles,
neural tabular models, and tabular foundation models.
We use each model's default settings without task-specific
hyperparameter tuning.
Comparisons use the supported datasets and splits
for each model.\footnote{TabPFN v2, TabPFN 2.5,
and TabFM exclude two MulTaBench text datasets
because of class-count limits; TabFM also lacks
paired results for eight STRABLE datasets.}
The evaluated models are listed in
Appendix~\ref{app:models}.}

\subsection{Evaluation}
\label{sec:setup_protocol}
\label{sec:setup_efficiency}
\newtext{Following the MulTaBench evaluation protocol,
we use five train/test splits per
dataset, with at most 10,000 training examples
per split. We evaluate classification using
AUC and regression using $R^2$. For each
dataset, we average paired Frozen--LightTAR
differences first across splits for each model
and then across supported models. We report the mean AUC gain across classification datasets and the mean
$R^2$ gain across regression datasets. We compute confidence intervals
by bootstrapping datasets within each group
\citep{demsar2006statistical,efron1979bootstrap}.}

\newtext{Following TabArena and BeyondArena \citep{purucker2026beyondiid},
we compare model--representation pairs within each benchmark using Elo ratings.
Appendix~\ref{app:elo} describes how we compute these ratings.}

\begin{table*}[!t]
\centering
\caption{\newtext{\textbf{Predictive performance improvements
of LightTAR over Frozen across 21 learners.}
Regression tasks are evaluated with $R^2$, and
classification tasks with AUC. Average gains are
reported in metric points (one point corresponds
to 0.01 in the respective metric), with 95\%
confidence intervals obtained through
dataset-level bootstrapping. W/T/L counts dataset-level wins, ties,
and losses relative to Frozen after averaging paired differences
across splits and learners. STRABLE results are reported
both without and with the gate described in
Section~\ref{sec:results_transfer} and Appendix~\ref{app:gate}.}}
\label{tab:regime_summary}
\small
\setlength{\tabcolsep}{4pt}
\newtext{%
\begin{tabular*}{\textwidth}{@{\extracolsep{\fill}}llcccc@{}}
\toprule
 & & \multicolumn{2}{c}{\textbf{Regression}} &
     \multicolumn{2}{c}{\textbf{Classification}} \\
\cmidrule(lr){3-4}\cmidrule(lr){5-6}
\textbf{Benchmark} & \textbf{Modality} &
\textbf{Gain [95\% CI]} & \textbf{W/T/L} &
\textbf{Gain [95\% CI]} & \textbf{W/T/L} \\
\midrule
MulTaBench & Text  & $+1.72\;[+0.66,+3.07]$ & $9/0/1$  & $+1.18\;[+0.37,+2.43]$ & $10/0/0$ \\
MulTaBench & Image & $+1.63\;[+0.26,+3.19]$ & $7/0/1$   & $+0.44\;[+0.16,+0.71]$ & $10/0/2$ \\
STRABLE (ungated) & Text  & $+0.29\;[+0.09,+0.51]$ & $47/0/29$ & $-0.26\;[-0.75,+0.09]$ & $17/0/15$ \\
STRABLE (with gate) & Text & $+0.34\;[+0.17,+0.55]$ & $43/6/27$ & $+0.08\;[-0.04,+0.19]$ & $18/4/10$ \\
\bottomrule
\end{tabular*}}
\end{table*}

\markoff

\markoff
\unreviewedon
\unreviewedon
\section{Results}
\label{sec:results}

\newtext{In this section, we examine the effect of
LightTAR on the predictive performance of various
tabular learners across the three benchmark suites, the
cost of using LightTAR compared with Frozen
vectors, and the resulting tradeoff between
predictive quality and runtime.}

\subsection{Predictive Quality}
\label{sec:results_mtb}

\newtext{Figure~\ref{fig:panel} presents the Elo
scores of the 21 tabular learners on MulTaBench
text and image, when the text or image features
are represented using either Frozen vectors or
LightTAR vectors. Table~\ref{tab:regime_summary}
summarizes the average predictive improvements
across the 21 learners, broken down by
benchmark and task type. Appendix~\ref{app:mtb_raw_results}
reports the Frozen and LightTAR scores for every
learner and MulTaBench dataset.}

\paragraph{\newtext{Target benchmarks: MulTaBench.}}
\newtext{Replacing Frozen representations with
LightTAR improves the predictive performance
of tabular learners on 19 of the 20 MulTaBench
text datasets, including all 10 classification
tasks and 9 of the 10 regression tasks.
The improvement extends across different model
families: each of the 21 learners achieves
a higher Elo rating with LightTAR than with
Frozen. Since the tabular learners are diverse
and held fixed, this consistency suggests that
the gains originate from the more useful shared
representation rather than from any particular
tabular learner architecture. The same method,
without changing the projection settings, also
improves performance on 17 of the 20 MulTaBench
image datasets: 10 of the 12 classification
tasks and 7 of the 8 regression tasks.}
\label{sec:results_img}
\newtext{Some image datasets also contain text columns. Appendix~\ref{app:ablations}
examines each modality separately in control experiments.}

\begin{figure*}[t]
\centering
\includegraphics[width=\linewidth]{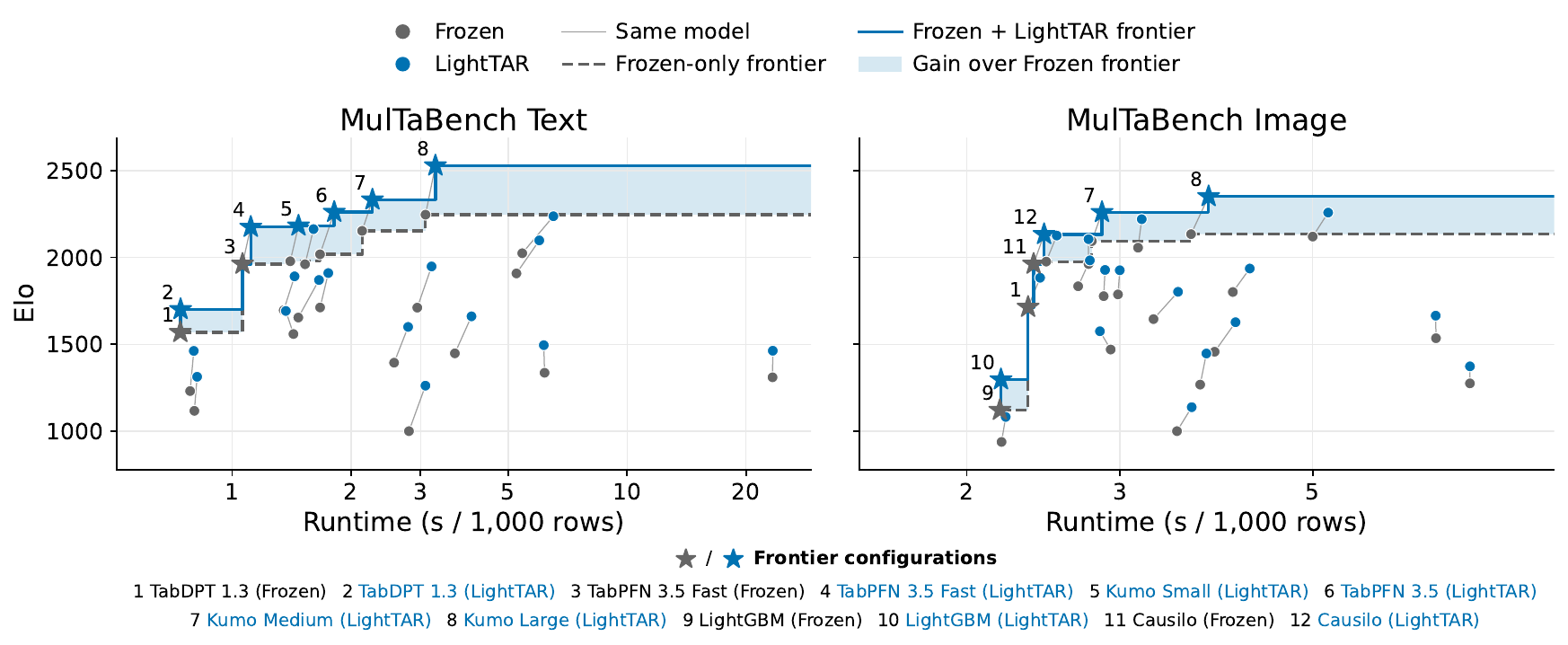}
\caption{\newtext{\textbf{Predictive quality versus runtime for Frozen
and LightTAR on MulTaBench text (left) and image (right).}
Each panel compares the models by their Elo ratings and runtime
in seconds per 1,000 train-and-test examples.
The runtime combines separately measured representation construction
and downstream learner fitting and prediction.}}
\label{fig:pareto}
\end{figure*}

\paragraph{\newtext{Results on STRABLE.}}
\label{sec:results_transfer}
\newtext{Unlike MulTaBench, STRABLE datasets were
not selected based on the predictive contribution
of text or the benefit of adapting its representation
to the task. LightTAR's results on this benchmark
are mixed: averaged across learners and datasets,
it improves regression performance but reduces
classification performance (Table~\ref{tab:regime_summary}).
We therefore evaluate a simple selection mechanism
(gate): cross-validation on the training data compares
the performance of a small LightGBM model with
Frozen and LightTAR representations, and the selected
representation is used to train the final tabular
learner. Gated LightTAR achieves a 48-point pooled
Elo gain over Frozen, with higher Elo ratings for
all 21 learners. Average predictive gains are positive
for both regression and classification.
Details of the selection mechanism, evaluation protocol,
and full results appear in Appendix~\ref{app:gate}.}

\subsection{Computational Efficiency}
\label{sec:results_cost}

\newtext{Using LightTAR introduces two sources
of additional computational cost relative to
Frozen: (1) constructing the additional representation,
and (2) representing each multimodal column
with more numerical features, which increases
the number of features processed by the tabular
learner and hence its runtime.}

\begin{table}[!b]
\centering
\caption{\newtext{\textbf{Total representation-construction
time on 20 MulTaBench text datasets.}}}
\label{tab:cost}
\small
\setlength{\tabcolsep}{4pt}
\begin{tabular*}{\columnwidth}{@{\extracolsep{\fill}}p{0.45\columnwidth}rr@{}}
\toprule
\textbf{Representation} & \textbf{Time (s)} &
\textbf{$\times$ LightTAR} \\
\midrule
Frozen (encoder + PCA) & $184.4$ & $0.994$ \\
LightTAR & $185.6$ & $1.00$ \\
Fine-Tuning with LoRA & $52{,}048.6$ & $280.5$ \\
\bottomrule
\end{tabular*}
\end{table}

\paragraph{\newtext{Runtime costs of representation construction.}}
\newtext{Table~\ref{tab:cost} compares the runtime cost
of constructing representations with Frozen
and LightTAR, as well as task-specific
fine-tuning of the encoder using LoRA,
following the protocol evaluated in the MulTaBench paper
(see Appendix~\ref{app:cost}).
Constructing LightTAR vectors adds only
\textbf{0.66\%} of the runtime required to compute Frozen
vectors, compared with an estimated
\textbf{28,130\%} increase for fine-tuning.
In both Frozen and LightTAR, most of the
cost comes from the encoder forward pass,
and the additional computations required
by LightTAR are negligible in comparison.}

\paragraph{\newtext{Runtime costs from additional features.}}
\newtext{Although LightTAR adds little time
to representation construction, the additional
features it produces also affect the fitting
and prediction time of the tabular learner.
For each of the 21 learners, we divide the runtime
with LightTAR by the runtime with Frozen.
For each representation, we measure representation construction and
tabular model fitting and prediction separately. Their sum gives the total runtime.
The median ratio shows that
LightTAR increases runtime by 5.9\% on text and
2.7\% on image relative to Frozen.}

\begin{table*}[!b]
\centering
\caption{\textbf{LightTAR ablations on MulTaBench text and image.}
Each entry reports the mean Elo gain over Frozen across five tabular learners,
expressed as a percentage of LightTAR's corresponding mean gain
(Frozen = 0\%; LightTAR = 100\%). Elo ratings are fitted jointly across
all learner--variant combinations, separately for classification and
regression within each benchmark. Negative values indicate a lower
mean Elo than Frozen. Layer descriptions refer to the input to the
additional projection or, for joint PCA, to the whole representation.
The shuffled-label control retains both layers, the PCA anchor,
regularization, and feature budget; only the projection labels are
permuted. The downstream learner uses the true labels.}
\label{tab:mechanism_ablations}
\small
\setlength{\tabcolsep}{4pt}
\begin{tabular*}{\textwidth}{@{\extracolsep{\fill}}p{0.46\textwidth}rr@{\hspace{12pt}}rr@{}}
\toprule
& \multicolumn{2}{c}{\textbf{MulTaBench Text}} &
  \multicolumn{2}{c}{\textbf{MulTaBench Image}} \\
\cmidrule(lr){2-3}\cmidrule(lr){4-5}
\textbf{Method} &
\textbf{Classification} & \textbf{Regression} &
\textbf{Classification} & \textbf{Regression} \\
\midrule
LightTAR: shuffled projection labels, two layers & $32.4\%$ & $42.6\%$ & $43.3\%$ & $48.3\%$ \\
Joint PCA: two layers, 35 components & $53.3\%$ & $73.6\%$ & $42.4\%$ & $41.1\%$ \\
LightTAR: final-layer directions only & $43.6\%$ & $48.2\%$ & $42.7\%$ & $56.9\%$ \\
LightTAR without PCA anchor (5 components) & $26.3\%$ & $27.2\%$ & $-71.9\%$ & $-96.7\%$ \\
PCA + PLS, last layer only & $26.2\%$ & $44.4\%$ & $32.5\%$ & $28.4\%$ \\
Textbook PLS: no PCA anchor, two layers, tuned depth & $44.3\%$ & $50.4\%$ & $28.0\%$ & $6.4\%$ \\
\bottomrule
\end{tabular*}
\end{table*}

\begin{table*}[t]
\centering
\caption{\textbf{Effect of text encoder size.}
Mean held-out AUC and $R^2$ across five tabular
learners using Frozen or LightTAR representations.
Gains are LightTAR minus Frozen for the same
encoder, in metric points (one point equals 0.01).}
\label{tab:encoder_scale}
\small
\setlength{\tabcolsep}{6pt}
\begin{tabular*}{\textwidth}{@{\extracolsep{\fill}}lrrrrrr@{}}
\toprule
& \multicolumn{3}{c}{\textbf{Classification (AUC)}} &
\multicolumn{3}{c}{\textbf{Regression ($R^2$)}} \\
\cmidrule(lr){2-4}\cmidrule(lr){5-7}
\textbf{Encoder} & \textbf{Frozen} & \textbf{LightTAR} & \textbf{Gain} &
\textbf{Frozen} & \textbf{LightTAR} & \textbf{Gain} \\
\midrule
\textit{e5-small-v2} & $0.9002$ & $0.9116$ & $+1.14$ &
$0.7244$ & $0.7423$ & $+1.79$ \\
\textit{e5-large-v2} & $0.9053$ & $0.9173$ & $+1.19$ &
$0.7289$ & $0.7456$ & $+1.68$ \\
\bottomrule
\end{tabular*}
\end{table*}

\paragraph{\newtext{Predictive quality versus computational cost.}}
\label{sec:pareto}
\newtext{Figure~\ref{fig:pareto} presents the
Elo rating of each learner on MulTaBench
text (left) and image (right) against its
runtime for Frozen (grey) and
LightTAR (blue).
For each benchmark, we compute two Pareto frontiers:
one among the 21 Frozen configurations and one
among all 42 Frozen and LightTAR configurations.
A configuration is included if no other configuration
has both a shorter runtime and a higher Elo rating.
The shaded region between the two curves shows the improvement
in attainable Elo at a fixed runtime budget when
LightTAR is available alongside Frozen.
Most configurations along the combined frontier use LightTAR.
To summarize the quality of the configuration sets
across both metrics, we use hypervolume~\citep{zitzler2000comparison},
which measures the area of the prediction-quality--runtime
space dominated by the configurations relative to a
common reference point. Comparing the hypervolume
attained by LightTAR with that of Frozen yields improvements
of \textbf{22.1\%} on text and
\textbf{16.3\%} on image.}

\markoff

\markoff
\unreviewedon
\unreviewedon
\section{Analysis}
\label{sec:analysis}

In this section, we examine the sources of LightTAR's
predictive gains, its robustness beyond MulTaBench,
and its sensitivity to projection settings and encoder
size. The controls in Table~\ref{tab:mechanism_ablations}
use a sample of five tabular learners (TabICLv2,
TabPFN 2.5, XGBoost, Kumo Large, and TabPFN 3.5),
using paired scores and task-specific Elo ratings.
Full details of the experimental setup and statistical analyses
appear in Appendix~\ref{app:ablations}.

\paragraph{What drives LightTAR's gains.}
Table~\ref{tab:mechanism_ablations} presents ablations
of LightTAR and comparisons with PCA- and PLS-based
alternatives. LightTAR achieves a larger mean Elo
gain over Frozen than every alternative in all
four benchmark--task groups.
\textbf{Using labels and the intermediate layer contributes to the gains.}
Shuffling the labels used to fit the projection,
while keeping the true labels for downstream training,
retains only 32.4--48.3\% of LightTAR's gain.
This control preserves both encoder layers, the PCA
anchor, regularization, and feature budget.
The remaining improvement may come from access to
the intermediate layer and the additional features.
However, joint PCA of the same two layers, matched
to LightTAR's feature budget, also yields a smaller
gain. Together, these comparisons support the contribution
of target-aware extraction beyond adding features
or accessing another layer. Complementing these results,
restricting LightTAR's directions to the final layer
retains 42.7--56.9\% of its gain, supporting the
contribution of intermediate-layer information.
\textbf{Combining these features with the PCA representation also matters.}
Removing the PCA anchor retains approximately one
quarter of the gain on text and yields lower Elo
than Frozen on image. Both textbook-PLS alternatives---with
an anchor and final-layer directions, or without
an anchor and with a tuned number of components
on two layers---also achieve smaller gains than
LightTAR in all four groups.

\paragraph{Sensitivity to encoder size and LightTAR parameters.}
Table~\ref{tab:encoder_scale} presents the
absolute predictive scores of the downstream
tabular learners and the improvements from
LightTAR over Frozen with \textit{e5-small-v2},
the encoder used in our main experiments with
approximately 33 million parameters, and its larger
variant, \textit{e5-large-v2}, with approximately
335 million parameters. The larger encoder improves absolute predictive performance.
LightTAR improves on Frozen by similar amounts with both encoders. Experiments in
Appendix~\ref{app:mwindow} examine how the gain over Frozen
varies with the number of features LightTAR adds ($s$)
and the regularization strength ($\gamma$). On MulTaBench
text, the gains are consistent and robust across a wide
range of values for both parameters. Additional experiments
in the same appendix show that removing regularization
does not significantly change performance on MulTaBench,
but reduces it on STRABLE.

\markoff

\markoff
\unreviewedon
\unreviewedon
\section{Discussion and Conclusion}
\label{sec:discussion}

We introduced LightTAR, a method that uses training labels when reducing
the dimensionality of text and image representations for tabular learners.
LightTAR augments the standard PCA representation with a few
features tailored to the prediction task. It learns these features using
a procedure inspired by PLS that combines information from the final and
intermediate layers of a frozen encoder. To reduce the risk of overfitting,
we use a regularization term that favors directions with high input variance
and moderates the influence of the training labels on the projection.

LightTAR improves predictive performance on datasets
with informative unstructured features, across a wide
range of tabular learners and with a small increase
in runtime. The controls support combining
target-aware directions, intermediate-layer
information, and PCA components for both text
and images, with the size of their contributions
varying across task types. When the benefit of
adapting these representations to the task is not
known in advance, the gate uses training data to
assess when LightTAR is worth applying.

We hope LightTAR will become a standard approach to building
multimodal representations for tabular learning. Constructing its
representations adds little computational cost, and the Pareto
analysis shows a favorable tradeoff between predictive quality and
runtime, even when the additional runtime of tabular learners is
taken into account. Moreover, LightTAR requires no changes to the
encoder or the downstream learner, making it easy to integrate into
existing multimodal tabular learning pipelines.

\paragraph{Limitations.}
LightTAR does not improve performance on every
dataset: its benefit depends on the predictive
information that linear projections can extract
from the selected layers of the frozen encoder.
Future work could extend LightTAR with nonlinear
projections, representations that account for
relationships between a text column and other
columns in the table, and additional simple
representations alongside frozen encoder features,
including interactions between these representations.
\markoff

\markoff

\bibliographystyle{icml2026}
\bibliography{references}

\appendix
\unreviewedon
\unreviewedon
\providecommand{\Sxx}{\widehat{S}_{xx}}
\providecommand{\Sxy}{\widehat{S}_{xy}}
\providecommand{\Lset}{\mathcal{L}}

\section{The LightTAR Algorithm}
\label{app:algo}

Algorithm~\ref{alg:lighttar} fits one projection per text or image
column. A single encoder pass gives the representations for all selected
layers. For text, we mean-pool the \textit{e5-small-v2} token representations
at each layer and normalize the resulting vector to unit length. For images,
we use the \textit{DINOv3-S} class token without further normalization.
Frozen fits PCA on centered final-layer embeddings. LightTAR also scales
each coordinate by its training standard deviation before fitting PCA.
For the supervised projection, we standardize the concatenated layer
representations and the regression target or, for classification, each class
indicator using the training rows.

\begin{algorithm}[!tbp]
\caption{Fitting LightTAR on layers $\Lset$ with anchor layer $\ell_a$,
$u$ PCA features, $s$ supervised features, and regularization $\gamma$.}
\label{alg:lighttar}
\begin{algorithmic}[1]
\STATE \textbf{fit}$\big(\{v_i,y_i\}_{i=1}^n;
                \Lset,\ell_a,u,s,\gamma\big)$:
\STATE Obtain the selected encoder representations
  $\phi^{(\ell)}(v_i)$ and concatenate them to
  $\vx_i$ (Eq.~\ref{eq:lighttar_main_layers}).
\STATE Standardize the anchor states and the concatenation
  using the training means and scales. Store these statistics.
\STATE $\mV_u\leftarrow$ top $u$ PCA directions
  of the standardized anchor matrix.
\STATE $\tilde{\mZ}^{(1)}\leftarrow\tilde{\mZ}$;\quad
  $\mY^{(1)}\leftarrow$ standardized targets.
\FOR{$j=1,\ldots,s$}
  \STATE $\Sxy^{(j)}\leftarrow
    n^{-1}\tilde{\mZ}^{(j)\top}\mY^{(j)}$;\quad
    $\Sxx^{(j)}\leftarrow
    n^{-1}\tilde{\mZ}^{(j)\top}\tilde{\mZ}^{(j)}$.
  \STATE $\vw_j\leftarrow$ leading eigenvector of
    $\Sxy^{(j)}\Sxy^{(j)\top}+\gamma\Sxx^{(j)}$.
  \STATE $\vt_j\leftarrow\tilde{\mZ}^{(j)}\vw_j$;\quad
    $\vp_j\leftarrow\tilde{\mZ}^{(j)\top}\vt_j/
      (\vt_j^\top\vt_j)$;\quad
    $\vq_j\leftarrow\mY^{(j)\top}\vt_j/(\vt_j^\top\vt_j)$.
  \STATE $\tilde{\mZ}^{(j+1)}\leftarrow
    \tilde{\mZ}^{(j)}-\vt_j\vp_j^\top$;\quad
    $\mY^{(j+1)}\leftarrow\mY^{(j)}-\vt_j\vq_j^\top$.
\ENDFOR
\STATE $\mW\leftarrow[\vw_1,\ldots,\vw_s]$;\quad
  $\mP\leftarrow[\vp_1,\ldots,\vp_s]$;\quad
  $\mR\leftarrow\mW(\mP^\top\mW)^{-1}$.
\STATE \textbf{transform}$(v)$:
  standardize the selected representations with the stored
  means and scales and return $\mathrm{L\text{-}TAR}(v)$ (Eq.~\ref{eq:lighttar_main_representation}).
\end{algorithmic}
\end{algorithm}

\section{Sensitivity to Projection Settings and Encoder Size}
\label{app:mwindow}
\paragraph{Number of supervised features and regularization strength.}
We vary the number of supervised features $s$ and the regularization
strength $\gamma=m/n$, where $m$ sets the regularization level and $n$
is the training set size. We selected
$m=60$ in preliminary experiments with TabICL on Fake Job Posting,
Jigsaw Toxicity, and Book Price. All three are included in the 20
MulTaBench text datasets evaluated here.
For each setting, we compare LightTAR with Frozen
on the same 290 learner--dataset--split combinations for
TabICLv2, TabPFN 2.5, and XGBoost. We average the paired score
differences over splits and then over learners within each dataset.
Across the 12 settings, the median dataset gain ranges from 0.0056
to 0.0100 on the original metric scale (Table~\ref{tab:kmgrid}).

\begin{table*}[!tbp]
\centering
\caption{\textbf{Sensitivity to $s$ and $m$ on MulTaBench text.}
Entries give the median gain over Frozen across 20 datasets, followed
by the number of datasets with a positive gain. Bold marks the setting
used in the main experiments.}
\label{tab:kmgrid}
\small
\begin{tabular}{lcccc}
\toprule
\textbf{Supervised features} & $m=20$ & $m=40$ & $m=60$ & $m=120$ \\
\midrule
$s=3$ & 0.0060 (19) & 0.0072 (19) & 0.0056 (18) & 0.0072 (17) \\
$s=5$ & 0.0059 (19) & 0.0089 (18) & \textbf{0.0094 (18)} & 0.0100 (18) \\
$s=8$ & 0.0080 (20) & 0.0081 (20) & 0.0094 (20) & 0.0073 (20) \\
\bottomrule
\end{tabular}
\end{table*}

\paragraph{Effect of regularization.}
We compare $\gamma=60/n$ with $\gamma=0$ using the five learners in
Section~\ref{sec:analysis}. Two-sided Wilcoxon tests on the dataset-level
score differences yield $p=1.00$ on MulTaBench text and $p=0.76$ on
image. On STRABLE, regularization improves performance on 77 of
108 datasets, with a median gain of 0.0014 ($p<10^{-8}$).

\paragraph{Encoder size.}
For Table~\ref{tab:encoder_scale}, we compare \textit{e5-small-v2} and
\textit{e5-large-v2} on the same 98 learner--dataset pairs and five
splits per pair. For each encoder, we average the held-out AUC or $R^2$
over splits, then over supported learners within each dataset, and
finally over datasets of the same task type. A paired two-sided Wilcoxon
test comparing the LightTAR gains over Frozen for the two encoders
across the 20 datasets gives $p=0.546$.

\section{The Evaluated Learners}
\label{app:models}

Table~\ref{tab:models} lists the 21 learners in Section~\ref{sec:setup_models}.
Spotify Genres and Wine Review have more classes than TabPFN v2, TabPFN 2.5,
and TabFM support. On eight STRABLE datasets, TabFM exceeded GPU memory
with LightTAR on all five splits, even with 95~GiB available. TabFM's paired
comparison with Frozen therefore covers 100 of the 108 datasets.

\begin{table}[!tbp]
\centering
\caption{\textbf{Tabular learners by model family.}}
\label{tab:models}
\small
\begin{tabular}{@{}lp{0.60\columnwidth}@{}}
\toprule
\textbf{Family} & \textbf{Learners} \\
\midrule
Tree ensembles & CatBoost, LightGBM, Random Forest, XGBoost \\
Neural networks & RealMLP, TabM \\
Foundation models & Causilo, EXAONE, Kumo Small, Medium and Large,
TabDPT 1.1 and 1.3, TabFM, TabICLv2, TabLDM, TabPFN v2, 2.5, v3,
3.5 and 3.5 Fast \\
\bottomrule
\end{tabular}
\end{table}

\section{Computing Elo Ratings}
\label{app:elo}

For the main comparison, we rate each of the 21 tabular learners with
each representation (Frozen and LightTAR), giving 42 combinations.
We fit a separate set of ratings for each benchmark. On each of the five
test splits of a dataset, we compare every pair of combinations for
which both scores are available. The higher score (AUC for classification,
$R^2$ for regression) earns one win. An exact tie gives each combination half a win.

We fit a Bradley--Terry model to these outcomes. If $r_i$ and $r_j$
are the Elo ratings of combinations $i$ and $j$, the model assigns $i$
a win probability of
\[
  \Pr(i \text{ beats } j) = \frac{1}{1+10^{(r_j-r_i)/400}}.
\]
We estimate the ratings by logistic regression with an $\ell_2$ penalty
and shift them by a common constant so that Random Forest with Frozen
has a rating of 1000 on each benchmark.

\section{Selection between Frozen and LightTAR}
\label{app:gate}

\paragraph{Selecting the representation.}
For each outer STRABLE split, we divide the training rows into five
inner folds, stratified by class for classification. Within each fold,
we fit the preprocessing and both projections on the inner training rows.
For each representation, we train LightGBM on the inner training rows,
including the structured features, and score the inner validation rows
by AUC or $R^2$. LightGBM uses 100 boosting rounds, learning rate 0.1,
maximum depth 3, at most seven leaves per tree, at least 20 training rows
per leaf, and seed 0.

We select LightTAR if it scores higher in at least three of the five
folds, and Frozen otherwise. All five folds were valid on each outer
split. The gate selected LightTAR on 408 of 540 splits. All final learners
share the decision for a given split. For each learner, we use its
held-out score for the selected representation fitted on the full outer
training set. This gives 11,300 paired learner--dataset--split observations.

\paragraph{Gate performance.}
We fit three pooled Elo ratings, one each for Frozen, LightTAR, and gated LightTAR.
Each comparison holds the learner, dataset, and outer split fixed. With Frozen anchored at 1000, LightTAR reaches 1033
(95\% CI $[1007,1058]$) and gated LightTAR reaches 1048 ($[1026,1068]$).
Confidence intervals use 1,000 bootstrap resamples of whole datasets.
Table~\ref{tab:regime_summary} reports the gains in AUC and $R^2$.
Comparing gated LightTAR with Frozen using a two-sided Wilcoxon
signed-rank test on the dataset-level differences gives $p=0.068$
for classification and $p=0.00064$ for regression.

\section{Full Results on MulTaBench}
\label{app:mtb_raw_results}

Tables~\ref{tab:mtb_raw_text_1}--\ref{tab:mtb_raw_text_7} report Frozen and
LightTAR scores for 414 learner--dataset pairs on MulTaBench text, and
Tables~\ref{tab:mtb_raw_image_1}--\ref{tab:mtb_raw_image_7} report scores for
420 pairs on MulTaBench image. Each entry gives the mean $\pm$ sample standard
deviation across the same five splits for both representations, using AUC for
classification and $R^2$ for regression. Dashes mark the class-count exclusions
described in Appendix~\ref{app:models}.

\begin{table*}[!tbp]
\centering
\caption{\textbf{MulTaBench text (1/7).}}
\label{tab:mtb_raw_text_1}
\small
\setlength{\tabcolsep}{3pt}
\begin{tabular*}{\textwidth}{@{\extracolsep{\fill}}lrrrrrr@{}}
\toprule
\textbf{Learner} & \multicolumn{2}{c}{\shortstack{\textbf{Fake Job Posting}\\(AUC)}} & \multicolumn{2}{c}{\shortstack{\textbf{Jigsaw Toxicity}\\(AUC)}} & \multicolumn{2}{c}{\shortstack{\textbf{Kickstarter Funding}\\(AUC)}} \\
\cmidrule(lr){2-3}\cmidrule(lr){4-5}\cmidrule(lr){6-7}
 & \textbf{Frozen} & \textbf{LightTAR} & \textbf{Frozen} & \textbf{LightTAR} & \textbf{Frozen} & \textbf{LightTAR} \\
\midrule
CatBoost & $0.906\,\pm\,0.016$ & $0.918\,\pm\,0.026$ & $0.818\,\pm\,0.025$ & $0.885\,\pm\,0.020$ & $0.729\,\pm\,0.010$ & $0.748\,\pm\,0.015$ \\
LightGBM & $0.909\,\pm\,0.020$ & $0.928\,\pm\,0.030$ & $0.798\,\pm\,0.023$ & $0.877\,\pm\,0.017$ & $0.715\,\pm\,0.013$ & $0.744\,\pm\,0.015$ \\
XGBoost & $0.901\,\pm\,0.012$ & $0.914\,\pm\,0.027$ & $0.760\,\pm\,0.031$ & $0.861\,\pm\,0.021$ & $0.691\,\pm\,0.011$ & $0.730\,\pm\,0.008$ \\
Random Forest & $0.893\,\pm\,0.032$ & $0.906\,\pm\,0.042$ & $0.745\,\pm\,0.037$ & $0.842\,\pm\,0.034$ & $0.697\,\pm\,0.011$ & $0.730\,\pm\,0.017$ \\
TabFM & $0.976\,\pm\,0.008$ & $0.976\,\pm\,0.008$ & $0.845\,\pm\,0.019$ & $0.895\,\pm\,0.022$ & $0.762\,\pm\,0.009$ & $0.770\,\pm\,0.015$ \\
RealMLP & $0.893\,\pm\,0.022$ & $0.911\,\pm\,0.033$ & $0.828\,\pm\,0.022$ & $0.887\,\pm\,0.024$ & $0.722\,\pm\,0.012$ & $0.740\,\pm\,0.015$ \\
TabM & $0.933\,\pm\,0.013$ & $0.936\,\pm\,0.023$ & $0.799\,\pm\,0.069$ & $0.885\,\pm\,0.020$ & $0.740\,\pm\,0.019$ & $0.748\,\pm\,0.016$ \\
TabICLv2 & $0.948\,\pm\,0.017$ & $0.956\,\pm\,0.016$ & $0.835\,\pm\,0.019$ & $0.891\,\pm\,0.022$ & $0.743\,\pm\,0.009$ & $0.759\,\pm\,0.012$ \\
TabDPT 1.1 & $0.945\,\pm\,0.017$ & $0.945\,\pm\,0.023$ & $0.801\,\pm\,0.025$ & $0.859\,\pm\,0.019$ & $0.695\,\pm\,0.015$ & $0.711\,\pm\,0.017$ \\
TabDPT 1.3 & $0.955\,\pm\,0.017$ & $0.960\,\pm\,0.016$ & $0.833\,\pm\,0.024$ & $0.892\,\pm\,0.022$ & $0.704\,\pm\,0.016$ & $0.728\,\pm\,0.016$ \\
TabPFN v2 & $0.913\,\pm\,0.022$ & $0.913\,\pm\,0.026$ & $0.823\,\pm\,0.024$ & $0.893\,\pm\,0.021$ & $0.725\,\pm\,0.013$ & $0.751\,\pm\,0.013$ \\
TabPFN 2.5 & $0.945\,\pm\,0.013$ & $0.953\,\pm\,0.016$ & $0.834\,\pm\,0.021$ & $0.892\,\pm\,0.023$ & $0.744\,\pm\,0.011$ & $0.758\,\pm\,0.013$ \\
TabPFN v3 & $0.957\,\pm\,0.017$ & $0.959\,\pm\,0.017$ & $0.836\,\pm\,0.021$ & $0.894\,\pm\,0.022$ & $0.746\,\pm\,0.010$ & $0.760\,\pm\,0.015$ \\
TabPFN 3.5 & $0.967\,\pm\,0.014$ & $0.973\,\pm\,0.012$ & $0.839\,\pm\,0.020$ & $0.894\,\pm\,0.022$ & $0.755\,\pm\,0.011$ & $0.763\,\pm\,0.014$ \\
TabPFN 3.5 Fast & $0.965\,\pm\,0.013$ & $0.972\,\pm\,0.012$ & $0.836\,\pm\,0.022$ & $0.893\,\pm\,0.022$ & $0.752\,\pm\,0.010$ & $0.763\,\pm\,0.013$ \\
Kumo Small & $0.966\,\pm\,0.012$ & $0.967\,\pm\,0.012$ & $0.838\,\pm\,0.020$ & $0.893\,\pm\,0.022$ & $0.751\,\pm\,0.011$ & $0.761\,\pm\,0.014$ \\
Kumo Medium & $0.973\,\pm\,0.010$ & $0.973\,\pm\,0.012$ & $0.839\,\pm\,0.019$ & $0.893\,\pm\,0.023$ & $0.760\,\pm\,0.009$ & $0.768\,\pm\,0.014$ \\
Kumo Large & $0.975\,\pm\,0.008$ & $0.977\,\pm\,0.010$ & $0.838\,\pm\,0.019$ & $0.895\,\pm\,0.022$ & $0.764\,\pm\,0.009$ & $0.771\,\pm\,0.014$ \\
Causilo & $0.967\,\pm\,0.010$ & $0.970\,\pm\,0.011$ & $0.839\,\pm\,0.020$ & $0.893\,\pm\,0.023$ & $0.753\,\pm\,0.012$ & $0.762\,\pm\,0.014$ \\
TabLDM & $0.947\,\pm\,0.018$ & $0.953\,\pm\,0.020$ & $0.832\,\pm\,0.019$ & $0.892\,\pm\,0.023$ & $0.745\,\pm\,0.010$ & $0.758\,\pm\,0.014$ \\
EXAONE & $0.967\,\pm\,0.013$ & $0.968\,\pm\,0.015$ & $0.835\,\pm\,0.022$ & $0.892\,\pm\,0.022$ & $0.749\,\pm\,0.011$ & $0.761\,\pm\,0.014$ \\
\bottomrule
\end{tabular*}
\par\medskip
\centering
\caption{\textbf{MulTaBench text (2/7).}}
\label{tab:mtb_raw_text_2}
\small
\setlength{\tabcolsep}{3pt}
\begin{tabular*}{\textwidth}{@{\extracolsep{\fill}}lrrrrrr@{}}
\toprule
\textbf{Learner} & \multicolumn{2}{c}{\shortstack{\textbf{Data Scientist Salary}\\(AUC)}} & \multicolumn{2}{c}{\shortstack{\textbf{Michelin Restaurants}\\(AUC)}} & \multicolumn{2}{c}{\shortstack{\textbf{Product Sentiment}\\(AUC)}} \\
\cmidrule(lr){2-3}\cmidrule(lr){4-5}\cmidrule(lr){6-7}
 & \textbf{Frozen} & \textbf{LightTAR} & \textbf{Frozen} & \textbf{LightTAR} & \textbf{Frozen} & \textbf{LightTAR} \\
\midrule
CatBoost & $0.820\,\pm\,0.004$ & $0.826\,\pm\,0.004$ & $0.905\,\pm\,0.003$ & $0.918\,\pm\,0.002$ & $0.900\,\pm\,0.013$ & $0.916\,\pm\,0.011$ \\
LightGBM & $0.800\,\pm\,0.003$ & $0.805\,\pm\,0.002$ & $0.898\,\pm\,0.003$ & $0.911\,\pm\,0.007$ & $0.872\,\pm\,0.022$ & $0.899\,\pm\,0.021$ \\
XGBoost & $0.806\,\pm\,0.006$ & $0.812\,\pm\,0.004$ & $0.877\,\pm\,0.006$ & $0.899\,\pm\,0.008$ & $0.895\,\pm\,0.017$ & $0.900\,\pm\,0.013$ \\
Random Forest & $0.775\,\pm\,0.004$ & $0.783\,\pm\,0.005$ & $0.858\,\pm\,0.011$ & $0.878\,\pm\,0.007$ & $0.870\,\pm\,0.015$ & $0.890\,\pm\,0.011$ \\
TabFM & $0.849\,\pm\,0.006$ & $0.854\,\pm\,0.006$ & $0.919\,\pm\,0.004$ & $0.931\,\pm\,0.004$ & $0.929\,\pm\,0.004$ & $0.930\,\pm\,0.008$ \\
RealMLP & $0.827\,\pm\,0.003$ & $0.832\,\pm\,0.004$ & $0.906\,\pm\,0.003$ & $0.916\,\pm\,0.002$ & $0.900\,\pm\,0.008$ & $0.910\,\pm\,0.011$ \\
TabM & $0.836\,\pm\,0.004$ & $0.838\,\pm\,0.005$ & $0.912\,\pm\,0.003$ & $0.923\,\pm\,0.004$ & $0.899\,\pm\,0.012$ & $0.914\,\pm\,0.011$ \\
TabICLv2 & $0.841\,\pm\,0.005$ & $0.844\,\pm\,0.004$ & $0.914\,\pm\,0.005$ & $0.924\,\pm\,0.004$ & $0.923\,\pm\,0.004$ & $0.926\,\pm\,0.008$ \\
TabDPT 1.1 & $0.852\,\pm\,0.005$ & $0.856\,\pm\,0.006$ & $0.854\,\pm\,0.006$ & $0.876\,\pm\,0.007$ & $0.917\,\pm\,0.006$ & $0.924\,\pm\,0.007$ \\
TabDPT 1.3 & $0.852\,\pm\,0.004$ & $0.856\,\pm\,0.005$ & $0.866\,\pm\,0.006$ & $0.886\,\pm\,0.006$ & $0.923\,\pm\,0.002$ & $0.928\,\pm\,0.007$ \\
TabPFN v2 & $0.830\,\pm\,0.005$ & $0.831\,\pm\,0.005$ & $0.908\,\pm\,0.006$ & $0.921\,\pm\,0.005$ & $0.915\,\pm\,0.005$ & $0.920\,\pm\,0.011$ \\
TabPFN 2.5 & $0.837\,\pm\,0.004$ & $0.840\,\pm\,0.005$ & $0.915\,\pm\,0.004$ & $0.927\,\pm\,0.005$ & $0.918\,\pm\,0.002$ & $0.923\,\pm\,0.011$ \\
TabPFN v3 & $0.826\,\pm\,0.004$ & $0.832\,\pm\,0.004$ & $0.911\,\pm\,0.004$ & $0.925\,\pm\,0.004$ & $0.921\,\pm\,0.004$ & $0.926\,\pm\,0.009$ \\
TabPFN 3.5 & $0.848\,\pm\,0.005$ & $0.852\,\pm\,0.005$ & $0.918\,\pm\,0.004$ & $0.930\,\pm\,0.004$ & $0.926\,\pm\,0.003$ & $0.928\,\pm\,0.009$ \\
TabPFN 3.5 Fast & $0.846\,\pm\,0.004$ & $0.848\,\pm\,0.004$ & $0.916\,\pm\,0.004$ & $0.927\,\pm\,0.004$ & $0.924\,\pm\,0.003$ & $0.928\,\pm\,0.010$ \\
Kumo Small & $0.843\,\pm\,0.004$ & $0.845\,\pm\,0.004$ & $0.917\,\pm\,0.004$ & $0.928\,\pm\,0.004$ & $0.926\,\pm\,0.003$ & $0.930\,\pm\,0.008$ \\
Kumo Medium & $0.848\,\pm\,0.006$ & $0.850\,\pm\,0.006$ & $0.918\,\pm\,0.003$ & $0.930\,\pm\,0.004$ & $0.927\,\pm\,0.005$ & $0.928\,\pm\,0.009$ \\
Kumo Large & $0.848\,\pm\,0.005$ & $0.853\,\pm\,0.006$ & $0.919\,\pm\,0.003$ & $0.931\,\pm\,0.003$ & $0.928\,\pm\,0.003$ & $0.930\,\pm\,0.008$ \\
Causilo & $0.848\,\pm\,0.004$ & $0.850\,\pm\,0.005$ & $0.916\,\pm\,0.005$ & $0.928\,\pm\,0.004$ & $0.923\,\pm\,0.003$ & $0.929\,\pm\,0.008$ \\
TabLDM & $0.837\,\pm\,0.004$ & $0.839\,\pm\,0.004$ & $0.913\,\pm\,0.004$ & $0.924\,\pm\,0.004$ & $0.920\,\pm\,0.005$ & $0.926\,\pm\,0.007$ \\
EXAONE & $0.839\,\pm\,0.005$ & $0.842\,\pm\,0.005$ & $0.915\,\pm\,0.004$ & $0.928\,\pm\,0.003$ & $0.921\,\pm\,0.007$ & $0.926\,\pm\,0.012$ \\
\bottomrule
\end{tabular*}
\end{table*}

\begin{table*}[!tbp]
\centering
\caption{\textbf{MulTaBench text (3/7).}}
\label{tab:mtb_raw_text_3}
\small
\setlength{\tabcolsep}{3pt}
\begin{tabular*}{\textwidth}{@{\extracolsep{\fill}}lrrrrrr@{}}
\toprule
\textbf{Learner} & \multicolumn{2}{c}{\shortstack{\textbf{Spotify Genres}\\(AUC)}} & \multicolumn{2}{c}{\shortstack{\textbf{US Accidents}\\(AUC)}} & \multicolumn{2}{c}{\shortstack{\textbf{Wine Review}\\(AUC)}} \\
\cmidrule(lr){2-3}\cmidrule(lr){4-5}\cmidrule(lr){6-7}
 & \textbf{Frozen} & \textbf{LightTAR} & \textbf{Frozen} & \textbf{LightTAR} & \textbf{Frozen} & \textbf{LightTAR} \\
\midrule
CatBoost & $0.960\,\pm\,0.002$ & $0.962\,\pm\,0.002$ & $0.971\,\pm\,0.005$ & $0.976\,\pm\,0.004$ & $0.975\,\pm\,0.001$ & $0.975\,\pm\,0.001$ \\
LightGBM & $0.897\,\pm\,0.003$ & $0.900\,\pm\,0.003$ & $0.966\,\pm\,0.005$ & $0.970\,\pm\,0.008$ & $0.965\,\pm\,0.001$ & $0.965\,\pm\,0.002$ \\
XGBoost & $0.931\,\pm\,0.002$ & $0.932\,\pm\,0.002$ & $0.968\,\pm\,0.004$ & $0.971\,\pm\,0.004$ & $0.964\,\pm\,0.001$ & $0.963\,\pm\,0.001$ \\
Random Forest & $0.918\,\pm\,0.005$ & $0.920\,\pm\,0.003$ & $0.950\,\pm\,0.014$ & $0.965\,\pm\,0.006$ & $0.952\,\pm\,0.005$ & $0.955\,\pm\,0.004$ \\
TabFM & -- & -- & $0.983\,\pm\,0.004$ & $0.984\,\pm\,0.004$ & -- & -- \\
RealMLP & $0.958\,\pm\,0.001$ & $0.960\,\pm\,0.002$ & $0.964\,\pm\,0.010$ & $0.966\,\pm\,0.011$ & $0.977\,\pm\,0.001$ & $0.978\,\pm\,0.001$ \\
TabM & $0.959\,\pm\,0.002$ & $0.962\,\pm\,0.001$ & $0.971\,\pm\,0.004$ & $0.968\,\pm\,0.013$ & $0.978\,\pm\,0.002$ & $0.978\,\pm\,0.002$ \\
TabICLv2 & $0.974\,\pm\,0.002$ & $0.973\,\pm\,0.001$ & $0.979\,\pm\,0.003$ & $0.981\,\pm\,0.004$ & $0.981\,\pm\,0.001$ & $0.980\,\pm\,0.001$ \\
TabDPT 1.1 & $0.861\,\pm\,0.005$ & $0.865\,\pm\,0.005$ & $0.931\,\pm\,0.006$ & $0.956\,\pm\,0.005$ & $0.940\,\pm\,0.005$ & $0.940\,\pm\,0.005$ \\
TabDPT 1.3 & $0.949\,\pm\,0.001$ & $0.951\,\pm\,0.002$ & $0.956\,\pm\,0.004$ & $0.967\,\pm\,0.005$ & $0.977\,\pm\,0.002$ & $0.977\,\pm\,0.002$ \\
TabPFN v2 & -- & -- & $0.976\,\pm\,0.003$ & $0.976\,\pm\,0.005$ & -- & -- \\
TabPFN 2.5 & -- & -- & $0.979\,\pm\,0.004$ & $0.981\,\pm\,0.004$ & -- & -- \\
TabPFN v3 & $0.976\,\pm\,0.001$ & $0.977\,\pm\,0.001$ & $0.978\,\pm\,0.004$ & $0.980\,\pm\,0.004$ & $0.981\,\pm\,0.001$ & $0.981\,\pm\,0.001$ \\
TabPFN 3.5 & $0.985\,\pm\,0.001$ & $0.985\,\pm\,0.001$ & $0.981\,\pm\,0.003$ & $0.982\,\pm\,0.004$ & $0.983\,\pm\,0.001$ & $0.983\,\pm\,0.001$ \\
TabPFN 3.5 Fast & $0.983\,\pm\,0.001$ & $0.984\,\pm\,0.001$ & $0.980\,\pm\,0.003$ & $0.981\,\pm\,0.004$ & $0.983\,\pm\,0.001$ & $0.983\,\pm\,0.001$ \\
Kumo Small & $0.980\,\pm\,0.001$ & $0.980\,\pm\,0.001$ & $0.981\,\pm\,0.004$ & $0.982\,\pm\,0.005$ & $0.983\,\pm\,0.001$ & $0.983\,\pm\,0.001$ \\
Kumo Medium & $0.983\,\pm\,0.000$ & $0.983\,\pm\,0.000$ & $0.983\,\pm\,0.004$ & $0.983\,\pm\,0.004$ & $0.983\,\pm\,0.001$ & $0.983\,\pm\,0.001$ \\
Kumo Large & $0.984\,\pm\,0.001$ & $0.984\,\pm\,0.001$ & $0.983\,\pm\,0.004$ & $0.984\,\pm\,0.004$ & $0.984\,\pm\,0.001$ & $0.984\,\pm\,0.001$ \\
Causilo & $0.978\,\pm\,0.001$ & $0.978\,\pm\,0.001$ & $0.981\,\pm\,0.003$ & $0.982\,\pm\,0.004$ & $0.982\,\pm\,0.001$ & $0.982\,\pm\,0.001$ \\
TabLDM & $0.975\,\pm\,0.001$ & $0.975\,\pm\,0.001$ & $0.979\,\pm\,0.003$ & $0.980\,\pm\,0.004$ & $0.980\,\pm\,0.001$ & $0.980\,\pm\,0.001$ \\
EXAONE & $0.978\,\pm\,0.001$ & $0.978\,\pm\,0.001$ & $0.979\,\pm\,0.004$ & $0.982\,\pm\,0.004$ & $0.981\,\pm\,0.001$ & $0.981\,\pm\,0.001$ \\
\bottomrule
\end{tabular*}
\par\medskip
\centering
\caption{\textbf{MulTaBench text (4/7).}}
\label{tab:mtb_raw_text_4}
\small
\setlength{\tabcolsep}{3pt}
\begin{tabular*}{\textwidth}{@{\extracolsep{\fill}}lrrrrrr@{}}
\toprule
\textbf{Learner} & \multicolumn{2}{c}{\shortstack{\textbf{Women Clothing Review}\\(AUC)}} & \multicolumn{2}{c}{\shortstack{\textbf{Babies Prices}\\($R^2$)}} & \multicolumn{2}{c}{\shortstack{\textbf{Book Price}\\($R^2$)}} \\
\cmidrule(lr){2-3}\cmidrule(lr){4-5}\cmidrule(lr){6-7}
 & \textbf{Frozen} & \textbf{LightTAR} & \textbf{Frozen} & \textbf{LightTAR} & \textbf{Frozen} & \textbf{LightTAR} \\
\midrule
CatBoost & $0.903\,\pm\,0.004$ & $0.905\,\pm\,0.004$ & $0.872\,\pm\,0.030$ & $0.877\,\pm\,0.025$ & $0.543\,\pm\,0.024$ & $0.580\,\pm\,0.026$ \\
LightGBM & $0.897\,\pm\,0.003$ & $0.899\,\pm\,0.004$ & $0.847\,\pm\,0.040$ & $0.849\,\pm\,0.033$ & $0.501\,\pm\,0.025$ & $0.565\,\pm\,0.015$ \\
XGBoost & $0.896\,\pm\,0.003$ & $0.899\,\pm\,0.003$ & $0.836\,\pm\,0.046$ & $0.832\,\pm\,0.037$ & $0.454\,\pm\,0.022$ & $0.506\,\pm\,0.032$ \\
Random Forest & $0.892\,\pm\,0.003$ & $0.897\,\pm\,0.005$ & $0.822\,\pm\,0.050$ & $0.839\,\pm\,0.024$ & $0.480\,\pm\,0.031$ & $0.541\,\pm\,0.033$ \\
TabFM & $0.911\,\pm\,0.003$ & $0.913\,\pm\,0.003$ & $0.941\,\pm\,0.015$ & $0.931\,\pm\,0.020$ & $0.610\,\pm\,0.025$ & $0.630\,\pm\,0.028$ \\
RealMLP & $0.906\,\pm\,0.003$ & $0.907\,\pm\,0.003$ & $0.856\,\pm\,0.050$ & $0.866\,\pm\,0.036$ & $0.533\,\pm\,0.027$ & $0.540\,\pm\,0.039$ \\
TabM & $0.907\,\pm\,0.003$ & $0.907\,\pm\,0.005$ & $0.900\,\pm\,0.031$ & $0.892\,\pm\,0.034$ & $0.560\,\pm\,0.024$ & $0.578\,\pm\,0.023$ \\
TabICLv2 & $0.908\,\pm\,0.004$ & $0.910\,\pm\,0.004$ & $0.932\,\pm\,0.019$ & $0.928\,\pm\,0.023$ & $0.592\,\pm\,0.036$ & $0.618\,\pm\,0.026$ \\
TabDPT 1.1 & $0.903\,\pm\,0.005$ & $0.906\,\pm\,0.004$ & $0.858\,\pm\,0.021$ & $0.869\,\pm\,0.026$ & $0.549\,\pm\,0.033$ & $0.575\,\pm\,0.028$ \\
TabDPT 1.3 & $0.909\,\pm\,0.004$ & $0.910\,\pm\,0.003$ & $0.918\,\pm\,0.022$ & $0.904\,\pm\,0.026$ & $0.579\,\pm\,0.031$ & $0.593\,\pm\,0.030$ \\
TabPFN v2 & $0.908\,\pm\,0.003$ & $0.910\,\pm\,0.003$ & $0.905\,\pm\,0.021$ & $0.896\,\pm\,0.018$ & $0.510\,\pm\,0.035$ & $0.545\,\pm\,0.032$ \\
TabPFN 2.5 & $0.909\,\pm\,0.004$ & $0.911\,\pm\,0.003$ & $0.914\,\pm\,0.020$ & $0.911\,\pm\,0.026$ & $0.570\,\pm\,0.033$ & $0.601\,\pm\,0.023$ \\
TabPFN v3 & $0.908\,\pm\,0.004$ & $0.909\,\pm\,0.004$ & $0.927\,\pm\,0.018$ & $0.924\,\pm\,0.022$ & $0.568\,\pm\,0.031$ & $0.603\,\pm\,0.029$ \\
TabPFN 3.5 & $0.910\,\pm\,0.003$ & $0.911\,\pm\,0.003$ & $0.938\,\pm\,0.018$ & $0.934\,\pm\,0.021$ & $0.616\,\pm\,0.025$ & $0.644\,\pm\,0.020$ \\
TabPFN 3.5 Fast & $0.910\,\pm\,0.003$ & $0.911\,\pm\,0.003$ & $0.939\,\pm\,0.016$ & $0.936\,\pm\,0.020$ & $0.617\,\pm\,0.020$ & $0.640\,\pm\,0.020$ \\
Kumo Small & $0.909\,\pm\,0.003$ & $0.911\,\pm\,0.003$ & $0.941\,\pm\,0.016$ & $0.934\,\pm\,0.021$ & $0.621\,\pm\,0.027$ & $0.639\,\pm\,0.023$ \\
Kumo Medium & $0.910\,\pm\,0.003$ & $0.911\,\pm\,0.003$ & $0.945\,\pm\,0.015$ & $0.934\,\pm\,0.021$ & $0.631\,\pm\,0.021$ & $0.649\,\pm\,0.016$ \\
Kumo Large & $0.911\,\pm\,0.003$ & $0.913\,\pm\,0.003$ & $0.946\,\pm\,0.018$ & $0.938\,\pm\,0.020$ & $0.640\,\pm\,0.019$ & $0.661\,\pm\,0.013$ \\
Causilo & $0.910\,\pm\,0.003$ & $0.911\,\pm\,0.003$ & $0.941\,\pm\,0.017$ & $0.934\,\pm\,0.022$ & $0.621\,\pm\,0.030$ & $0.641\,\pm\,0.031$ \\
TabLDM & $0.907\,\pm\,0.003$ & $0.910\,\pm\,0.003$ & $0.939\,\pm\,0.014$ & $0.930\,\pm\,0.021$ & $0.599\,\pm\,0.030$ & $0.620\,\pm\,0.024$ \\
EXAONE & $0.910\,\pm\,0.003$ & $0.911\,\pm\,0.003$ & $0.928\,\pm\,0.017$ & $0.918\,\pm\,0.024$ & $0.616\,\pm\,0.022$ & $0.633\,\pm\,0.014$ \\
\bottomrule
\end{tabular*}
\end{table*}

\begin{table*}[!tbp]
\centering
\caption{\textbf{MulTaBench text (5/7).}}
\label{tab:mtb_raw_text_5}
\small
\setlength{\tabcolsep}{3pt}
\begin{tabular*}{\textwidth}{@{\extracolsep{\fill}}lrrrrrr@{}}
\toprule
\textbf{Learner} & \multicolumn{2}{c}{\shortstack{\textbf{Book Readability}\\($R^2$)}} & \multicolumn{2}{c}{\shortstack{\textbf{Mercari Marketplace}\\($R^2$)}} & \multicolumn{2}{c}{\shortstack{\textbf{Montgomery Salaries}\\($R^2$)}} \\
\cmidrule(lr){2-3}\cmidrule(lr){4-5}\cmidrule(lr){6-7}
 & \textbf{Frozen} & \textbf{LightTAR} & \textbf{Frozen} & \textbf{LightTAR} & \textbf{Frozen} & \textbf{LightTAR} \\
\midrule
CatBoost & $0.819\,\pm\,0.021$ & $0.832\,\pm\,0.025$ & $0.436\,\pm\,0.024$ & $0.459\,\pm\,0.026$ & $0.972\,\pm\,0.011$ & $0.973\,\pm\,0.011$ \\
LightGBM & $0.810\,\pm\,0.020$ & $0.822\,\pm\,0.030$ & $0.409\,\pm\,0.026$ & $0.445\,\pm\,0.030$ & $0.969\,\pm\,0.012$ & $0.972\,\pm\,0.012$ \\
XGBoost & $0.784\,\pm\,0.029$ & $0.797\,\pm\,0.031$ & $0.354\,\pm\,0.019$ & $0.390\,\pm\,0.034$ & $0.963\,\pm\,0.014$ & $0.967\,\pm\,0.011$ \\
Random Forest & $0.788\,\pm\,0.024$ & $0.796\,\pm\,0.028$ & $0.363\,\pm\,0.024$ & $0.401\,\pm\,0.029$ & $0.966\,\pm\,0.014$ & $0.966\,\pm\,0.013$ \\
TabFM & $0.836\,\pm\,0.018$ & $0.850\,\pm\,0.024$ & $0.491\,\pm\,0.033$ & $0.504\,\pm\,0.036$ & $0.983\,\pm\,0.008$ & $0.985\,\pm\,0.006$ \\
RealMLP & $0.802\,\pm\,0.023$ & $0.814\,\pm\,0.020$ & $0.368\,\pm\,0.037$ & $0.413\,\pm\,0.034$ & $0.971\,\pm\,0.013$ & $0.971\,\pm\,0.010$ \\
TabM & $0.816\,\pm\,0.019$ & $0.831\,\pm\,0.019$ & $0.454\,\pm\,0.028$ & $0.481\,\pm\,0.028$ & $0.975\,\pm\,0.013$ & $0.976\,\pm\,0.012$ \\
TabICLv2 & $0.830\,\pm\,0.018$ & $0.851\,\pm\,0.021$ & $0.441\,\pm\,0.031$ & $0.456\,\pm\,0.029$ & $0.977\,\pm\,0.009$ & $0.978\,\pm\,0.009$ \\
TabDPT 1.1 & $0.792\,\pm\,0.018$ & $0.796\,\pm\,0.022$ & $0.449\,\pm\,0.023$ & $0.464\,\pm\,0.032$ & $0.930\,\pm\,0.013$ & $0.921\,\pm\,0.008$ \\
TabDPT 1.3 & $0.798\,\pm\,0.022$ & $0.807\,\pm\,0.023$ & $0.461\,\pm\,0.023$ & $0.477\,\pm\,0.029$ & $0.977\,\pm\,0.011$ & $0.974\,\pm\,0.008$ \\
TabPFN v2 & $0.827\,\pm\,0.015$ & $0.843\,\pm\,0.018$ & $0.417\,\pm\,0.024$ & $0.454\,\pm\,0.028$ & $0.974\,\pm\,0.012$ & $0.976\,\pm\,0.011$ \\
TabPFN 2.5 & $0.832\,\pm\,0.017$ & $0.850\,\pm\,0.019$ & $0.473\,\pm\,0.029$ & $0.490\,\pm\,0.029$ & $0.979\,\pm\,0.012$ & $0.981\,\pm\,0.009$ \\
TabPFN v3 & $0.827\,\pm\,0.014$ & $0.846\,\pm\,0.018$ & $0.447\,\pm\,0.025$ & $0.467\,\pm\,0.031$ & $0.977\,\pm\,0.009$ & $0.979\,\pm\,0.008$ \\
TabPFN 3.5 & $0.834\,\pm\,0.019$ & $0.851\,\pm\,0.023$ & $0.481\,\pm\,0.030$ & $0.498\,\pm\,0.033$ & $0.984\,\pm\,0.010$ & $0.984\,\pm\,0.008$ \\
TabPFN 3.5 Fast & $0.833\,\pm\,0.019$ & $0.851\,\pm\,0.025$ & $0.484\,\pm\,0.034$ & $0.496\,\pm\,0.032$ & $0.983\,\pm\,0.009$ & $0.984\,\pm\,0.009$ \\
Kumo Small & $0.840\,\pm\,0.018$ & $0.854\,\pm\,0.021$ & $0.492\,\pm\,0.031$ & $0.499\,\pm\,0.034$ & $0.983\,\pm\,0.010$ & $0.982\,\pm\,0.010$ \\
Kumo Medium & $0.841\,\pm\,0.017$ & $0.855\,\pm\,0.020$ & $0.507\,\pm\,0.030$ & $0.515\,\pm\,0.029$ & $0.983\,\pm\,0.010$ & $0.983\,\pm\,0.008$ \\
Kumo Large & $0.844\,\pm\,0.017$ & $0.858\,\pm\,0.019$ & $0.513\,\pm\,0.030$ & $0.519\,\pm\,0.031$ & $0.984\,\pm\,0.010$ & $0.986\,\pm\,0.008$ \\
Causilo & $0.834\,\pm\,0.018$ & $0.849\,\pm\,0.024$ & $0.495\,\pm\,0.031$ & $0.503\,\pm\,0.033$ & $0.983\,\pm\,0.008$ & $0.983\,\pm\,0.007$ \\
TabLDM & $0.829\,\pm\,0.019$ & $0.852\,\pm\,0.020$ & $0.462\,\pm\,0.032$ & $0.479\,\pm\,0.032$ & $0.979\,\pm\,0.010$ & $0.978\,\pm\,0.011$ \\
EXAONE & $0.838\,\pm\,0.016$ & $0.854\,\pm\,0.020$ & $0.489\,\pm\,0.033$ & $0.497\,\pm\,0.033$ & $0.981\,\pm\,0.010$ & $0.982\,\pm\,0.009$ \\
\bottomrule
\end{tabular*}
\par\medskip
\centering
\caption{\textbf{MulTaBench text (6/7).}}
\label{tab:mtb_raw_text_6}
\small
\setlength{\tabcolsep}{3pt}
\begin{tabular*}{\textwidth}{@{\extracolsep{\fill}}lrrrrrr@{}}
\toprule
\textbf{Learner} & \multicolumn{2}{c}{\shortstack{\textbf{Rotten Tomatoes}\\($R^2$)}} & \multicolumn{2}{c}{\shortstack{\textbf{ScimagoJR Impact}\\($R^2$)}} & \multicolumn{2}{c}{\shortstack{\textbf{Vancouver Salaries}\\($R^2$)}} \\
\cmidrule(lr){2-3}\cmidrule(lr){4-5}\cmidrule(lr){6-7}
 & \textbf{Frozen} & \textbf{LightTAR} & \textbf{Frozen} & \textbf{LightTAR} & \textbf{Frozen} & \textbf{LightTAR} \\
\midrule
CatBoost & $0.508\,\pm\,0.039$ & $0.530\,\pm\,0.031$ & $0.868\,\pm\,0.012$ & $0.872\,\pm\,0.013$ & $0.764\,\pm\,0.021$ & $0.780\,\pm\,0.021$ \\
LightGBM & $0.477\,\pm\,0.036$ & $0.512\,\pm\,0.028$ & $0.860\,\pm\,0.011$ & $0.867\,\pm\,0.010$ & $0.741\,\pm\,0.019$ & $0.758\,\pm\,0.020$ \\
XGBoost & $0.426\,\pm\,0.041$ & $0.469\,\pm\,0.031$ & $0.833\,\pm\,0.015$ & $0.844\,\pm\,0.013$ & $0.725\,\pm\,0.023$ & $0.751\,\pm\,0.024$ \\
Random Forest & $0.437\,\pm\,0.035$ & $0.473\,\pm\,0.022$ & $0.830\,\pm\,0.018$ & $0.841\,\pm\,0.018$ & $0.717\,\pm\,0.025$ & $0.751\,\pm\,0.024$ \\
TabFM & $0.546\,\pm\,0.034$ & $0.575\,\pm\,0.044$ & $0.894\,\pm\,0.009$ & $0.895\,\pm\,0.005$ & $0.793\,\pm\,0.021$ & $0.801\,\pm\,0.022$ \\
RealMLP & $0.494\,\pm\,0.051$ & $0.531\,\pm\,0.027$ & $0.839\,\pm\,0.018$ & $0.829\,\pm\,0.047$ & $0.735\,\pm\,0.023$ & $0.738\,\pm\,0.022$ \\
TabM & $0.531\,\pm\,0.041$ & $0.546\,\pm\,0.035$ & $0.879\,\pm\,0.010$ & $0.872\,\pm\,0.016$ & $0.766\,\pm\,0.014$ & $0.768\,\pm\,0.018$ \\
TabICLv2 & $0.515\,\pm\,0.038$ & $0.533\,\pm\,0.039$ & $0.869\,\pm\,0.011$ & $0.876\,\pm\,0.012$ & $0.786\,\pm\,0.014$ & $0.795\,\pm\,0.018$ \\
TabDPT 1.1 & $0.414\,\pm\,0.022$ & $0.470\,\pm\,0.020$ & $0.790\,\pm\,0.021$ & $0.823\,\pm\,0.017$ & $0.769\,\pm\,0.016$ & $0.776\,\pm\,0.015$ \\
TabDPT 1.3 & $0.463\,\pm\,0.032$ & $0.501\,\pm\,0.031$ & $0.831\,\pm\,0.011$ & $0.848\,\pm\,0.016$ & $0.800\,\pm\,0.018$ & $0.801\,\pm\,0.018$ \\
TabPFN v2 & $0.537\,\pm\,0.035$ & $0.559\,\pm\,0.032$ & $0.878\,\pm\,0.011$ & $0.877\,\pm\,0.012$ & $0.755\,\pm\,0.019$ & $0.769\,\pm\,0.022$ \\
TabPFN 2.5 & $0.556\,\pm\,0.038$ & $0.578\,\pm\,0.035$ & $0.880\,\pm\,0.031$ & $0.889\,\pm\,0.009$ & $0.792\,\pm\,0.018$ & $0.795\,\pm\,0.020$ \\
TabPFN v3 & $0.537\,\pm\,0.030$ & $0.557\,\pm\,0.039$ & $0.879\,\pm\,0.006$ & $0.884\,\pm\,0.006$ & $0.784\,\pm\,0.016$ & $0.790\,\pm\,0.023$ \\
TabPFN 3.5 & $0.549\,\pm\,0.035$ & $0.575\,\pm\,0.039$ & $0.895\,\pm\,0.011$ & $0.897\,\pm\,0.006$ & $0.798\,\pm\,0.018$ & $0.804\,\pm\,0.020$ \\
TabPFN 3.5 Fast & $0.547\,\pm\,0.034$ & $0.571\,\pm\,0.041$ & $0.892\,\pm\,0.009$ & $0.891\,\pm\,0.006$ & $0.804\,\pm\,0.018$ & $0.810\,\pm\,0.022$ \\
Kumo Small & $0.561\,\pm\,0.039$ & $0.577\,\pm\,0.039$ & $0.897\,\pm\,0.010$ & $0.898\,\pm\,0.006$ & $0.798\,\pm\,0.019$ & $0.810\,\pm\,0.022$ \\
Kumo Medium & $0.579\,\pm\,0.038$ & $0.593\,\pm\,0.037$ & $0.904\,\pm\,0.011$ & $0.903\,\pm\,0.008$ & $0.821\,\pm\,0.021$ & $0.827\,\pm\,0.023$ \\
Kumo Large & $0.585\,\pm\,0.039$ & $0.598\,\pm\,0.037$ & $0.905\,\pm\,0.014$ & $0.908\,\pm\,0.007$ & $0.825\,\pm\,0.021$ & $0.829\,\pm\,0.021$ \\
Causilo & $0.549\,\pm\,0.036$ & $0.573\,\pm\,0.038$ & $0.888\,\pm\,0.008$ & $0.891\,\pm\,0.010$ & $0.798\,\pm\,0.020$ & $0.803\,\pm\,0.021$ \\
TabLDM & $0.509\,\pm\,0.038$ & $0.531\,\pm\,0.039$ & $0.874\,\pm\,0.011$ & $0.873\,\pm\,0.017$ & $0.794\,\pm\,0.019$ & $0.801\,\pm\,0.020$ \\
EXAONE & $0.565\,\pm\,0.038$ & $0.581\,\pm\,0.039$ & $0.899\,\pm\,0.006$ & $0.901\,\pm\,0.004$ & $0.801\,\pm\,0.018$ & $0.803\,\pm\,0.019$ \\
\bottomrule
\end{tabular*}
\end{table*}

\begin{table*}[!tbp]
\centering
\caption{\textbf{MulTaBench text (7/7).}}
\label{tab:mtb_raw_text_7}
\small
\setlength{\tabcolsep}{3pt}
\begin{tabular*}{\textwidth}{@{\extracolsep{\fill}}lrrrr@{}}
\toprule
\textbf{Learner} & \multicolumn{2}{c}{\shortstack{\textbf{Video Games Sales}\\($R^2$)}} & \multicolumn{2}{c}{\shortstack{\textbf{Zomato Restaurants}\\($R^2$)}} \\
\cmidrule(lr){2-3}\cmidrule(lr){4-5}
 & \textbf{Frozen} & \textbf{LightTAR} & \textbf{Frozen} & \textbf{LightTAR} \\
\midrule
CatBoost & $0.325\,\pm\,0.142$ & $0.447\,\pm\,0.081$ & $0.825\,\pm\,0.004$ & $0.832\,\pm\,0.003$ \\
LightGBM & $0.358\,\pm\,0.093$ & $0.447\,\pm\,0.105$ & $0.778\,\pm\,0.006$ & $0.790\,\pm\,0.006$ \\
XGBoost & $0.223\,\pm\,0.137$ & $0.356\,\pm\,0.142$ & $0.807\,\pm\,0.008$ & $0.821\,\pm\,0.007$ \\
Random Forest & $0.257\,\pm\,0.086$ & $0.392\,\pm\,0.093$ & $0.783\,\pm\,0.009$ & $0.789\,\pm\,0.004$ \\
TabFM & $0.549\,\pm\,0.103$ & $0.574\,\pm\,0.106$ & $0.864\,\pm\,0.001$ & $0.862\,\pm\,0.005$ \\
RealMLP & $0.354\,\pm\,0.135$ & $0.420\,\pm\,0.104$ & $0.842\,\pm\,0.007$ & $0.845\,\pm\,0.007$ \\
TabM & $0.485\,\pm\,0.138$ & $0.566\,\pm\,0.095$ & $0.844\,\pm\,0.006$ & $0.846\,\pm\,0.004$ \\
TabICLv2 & $0.443\,\pm\,0.091$ & $0.493\,\pm\,0.111$ & $0.872\,\pm\,0.001$ & $0.873\,\pm\,0.004$ \\
TabDPT 1.1 & $0.336\,\pm\,0.068$ & $0.377\,\pm\,0.073$ & $0.828\,\pm\,0.008$ & $0.839\,\pm\,0.007$ \\
TabDPT 1.3 & $0.398\,\pm\,0.086$ & $0.432\,\pm\,0.096$ & $0.853\,\pm\,0.007$ & $0.860\,\pm\,0.007$ \\
TabPFN v2 & $0.290\,\pm\,0.056$ & $0.317\,\pm\,0.056$ & $0.716\,\pm\,0.011$ & $0.722\,\pm\,0.012$ \\
TabPFN 2.5 & $0.406\,\pm\,0.066$ & $0.489\,\pm\,0.077$ & $0.865\,\pm\,0.006$ & $0.869\,\pm\,0.005$ \\
TabPFN v3 & $0.376\,\pm\,0.063$ & $0.438\,\pm\,0.084$ & $0.862\,\pm\,0.002$ & $0.865\,\pm\,0.004$ \\
TabPFN 3.5 & $0.561\,\pm\,0.104$ & $0.609\,\pm\,0.093$ & $0.881\,\pm\,0.003$ & $0.883\,\pm\,0.003$ \\
TabPFN 3.5 Fast & $0.514\,\pm\,0.102$ & $0.584\,\pm\,0.118$ & $0.878\,\pm\,0.004$ & $0.878\,\pm\,0.006$ \\
Kumo Small & $0.586\,\pm\,0.106$ & $0.642\,\pm\,0.088$ & $0.877\,\pm\,0.003$ & $0.879\,\pm\,0.003$ \\
Kumo Medium & $0.575\,\pm\,0.118$ & $0.641\,\pm\,0.119$ & $0.890\,\pm\,0.008$ & $0.892\,\pm\,0.006$ \\
Kumo Large & $0.618\,\pm\,0.117$ & $0.659\,\pm\,0.115$ & $0.890\,\pm\,0.008$ & $0.893\,\pm\,0.005$ \\
Causilo & $0.539\,\pm\,0.104$ & $0.612\,\pm\,0.118$ & $0.877\,\pm\,0.002$ & $0.878\,\pm\,0.003$ \\
TabLDM & $0.438\,\pm\,0.092$ & $0.524\,\pm\,0.133$ & $0.873\,\pm\,0.004$ & $0.876\,\pm\,0.003$ \\
EXAONE & $0.542\,\pm\,0.094$ & $0.595\,\pm\,0.081$ & $0.877\,\pm\,0.005$ & $0.876\,\pm\,0.004$ \\
\bottomrule
\end{tabular*}
\par\medskip
\centering
\caption{\textbf{MulTaBench image (1/7).}}
\label{tab:mtb_raw_image_1}
\small
\setlength{\tabcolsep}{3pt}
\begin{tabular*}{\textwidth}{@{\extracolsep{\fill}}lrrrrrr@{}}
\toprule
\textbf{Learner} & \multicolumn{2}{c}{\shortstack{\textbf{Celeb Attractiveness}\\(AUC)}} & \multicolumn{2}{c}{\shortstack{\textbf{Hateful Meme}\\(AUC)}} & \multicolumn{2}{c}{\shortstack{\textbf{Mammography CMMD}\\(AUC)}} \\
\cmidrule(lr){2-3}\cmidrule(lr){4-5}\cmidrule(lr){6-7}
 & \textbf{Frozen} & \textbf{LightTAR} & \textbf{Frozen} & \textbf{LightTAR} & \textbf{Frozen} & \textbf{LightTAR} \\
\midrule
CatBoost & $0.904\,\pm\,0.003$ & $0.911\,\pm\,0.004$ & $0.739\,\pm\,0.013$ & $0.743\,\pm\,0.011$ & $0.757\,\pm\,0.017$ & $0.771\,\pm\,0.016$ \\
LightGBM & $0.898\,\pm\,0.004$ & $0.906\,\pm\,0.006$ & $0.726\,\pm\,0.015$ & $0.726\,\pm\,0.014$ & $0.741\,\pm\,0.007$ & $0.754\,\pm\,0.014$ \\
XGBoost & $0.891\,\pm\,0.004$ & $0.901\,\pm\,0.006$ & $0.705\,\pm\,0.011$ & $0.712\,\pm\,0.014$ & $0.722\,\pm\,0.016$ & $0.722\,\pm\,0.008$ \\
Random Forest & $0.891\,\pm\,0.005$ & $0.901\,\pm\,0.005$ & $0.704\,\pm\,0.010$ & $0.700\,\pm\,0.005$ & $0.726\,\pm\,0.029$ & $0.737\,\pm\,0.026$ \\
TabFM & $0.911\,\pm\,0.006$ & $0.914\,\pm\,0.005$ & $0.819\,\pm\,0.013$ & $0.814\,\pm\,0.011$ & $0.926\,\pm\,0.008$ & $0.942\,\pm\,0.010$ \\
RealMLP & $0.907\,\pm\,0.006$ & $0.910\,\pm\,0.003$ & $0.752\,\pm\,0.036$ & $0.762\,\pm\,0.022$ & $0.760\,\pm\,0.016$ & $0.773\,\pm\,0.026$ \\
TabM & $0.908\,\pm\,0.005$ & $0.912\,\pm\,0.005$ & $0.775\,\pm\,0.010$ & $0.770\,\pm\,0.005$ & $0.745\,\pm\,0.028$ & $0.770\,\pm\,0.010$ \\
TabICLv2 & $0.909\,\pm\,0.006$ & $0.913\,\pm\,0.005$ & $0.772\,\pm\,0.013$ & $0.770\,\pm\,0.011$ & $0.861\,\pm\,0.007$ & $0.856\,\pm\,0.012$ \\
TabDPT 1.1 & $0.906\,\pm\,0.004$ & $0.909\,\pm\,0.004$ & $0.757\,\pm\,0.018$ & $0.758\,\pm\,0.015$ & $0.788\,\pm\,0.023$ & $0.785\,\pm\,0.022$ \\
TabDPT 1.3 & $0.909\,\pm\,0.006$ & $0.913\,\pm\,0.006$ & $0.766\,\pm\,0.012$ & $0.766\,\pm\,0.012$ & $0.796\,\pm\,0.020$ & $0.794\,\pm\,0.018$ \\
TabPFN v2 & $0.908\,\pm\,0.005$ & $0.913\,\pm\,0.005$ & $0.758\,\pm\,0.008$ & $0.759\,\pm\,0.010$ & $0.802\,\pm\,0.014$ & $0.805\,\pm\,0.013$ \\
TabPFN 2.5 & $0.909\,\pm\,0.006$ & $0.914\,\pm\,0.005$ & $0.769\,\pm\,0.011$ & $0.769\,\pm\,0.010$ & $0.806\,\pm\,0.018$ & $0.808\,\pm\,0.017$ \\
TabPFN v3 & $0.910\,\pm\,0.006$ & $0.914\,\pm\,0.005$ & $0.776\,\pm\,0.012$ & $0.773\,\pm\,0.012$ & $0.838\,\pm\,0.013$ & $0.839\,\pm\,0.016$ \\
TabPFN 3.5 & $0.911\,\pm\,0.006$ & $0.914\,\pm\,0.005$ & $0.786\,\pm\,0.015$ & $0.785\,\pm\,0.014$ & $0.872\,\pm\,0.010$ & $0.876\,\pm\,0.012$ \\
TabPFN 3.5 Fast & $0.910\,\pm\,0.006$ & $0.914\,\pm\,0.005$ & $0.786\,\pm\,0.012$ & $0.785\,\pm\,0.014$ & $0.861\,\pm\,0.008$ & $0.871\,\pm\,0.008$ \\
Kumo Small & $0.910\,\pm\,0.006$ & $0.914\,\pm\,0.005$ & $0.782\,\pm\,0.015$ & $0.770\,\pm\,0.014$ & $0.866\,\pm\,0.010$ & $0.864\,\pm\,0.013$ \\
Kumo Medium & $0.910\,\pm\,0.005$ & $0.914\,\pm\,0.005$ & $0.832\,\pm\,0.015$ & $0.815\,\pm\,0.011$ & $0.876\,\pm\,0.010$ & $0.880\,\pm\,0.010$ \\
Kumo Large & $0.910\,\pm\,0.005$ & $0.914\,\pm\,0.005$ & $0.805\,\pm\,0.015$ & $0.799\,\pm\,0.013$ & $0.883\,\pm\,0.015$ & $0.911\,\pm\,0.014$ \\
Causilo & $0.910\,\pm\,0.005$ & $0.915\,\pm\,0.005$ & $0.774\,\pm\,0.015$ & $0.777\,\pm\,0.014$ & $0.858\,\pm\,0.011$ & $0.853\,\pm\,0.010$ \\
TabLDM & $0.909\,\pm\,0.006$ & $0.914\,\pm\,0.005$ & $0.768\,\pm\,0.011$ & $0.765\,\pm\,0.011$ & $0.783\,\pm\,0.022$ & $0.777\,\pm\,0.018$ \\
EXAONE & $0.909\,\pm\,0.005$ & $0.914\,\pm\,0.006$ & $0.761\,\pm\,0.015$ & $0.747\,\pm\,0.014$ & $0.810\,\pm\,0.027$ & $0.811\,\pm\,0.015$ \\
\bottomrule
\end{tabular*}
\end{table*}

\begin{table*}[!tbp]
\centering
\caption{\textbf{MulTaBench image (2/7).}}
\label{tab:mtb_raw_image_2}
\small
\setlength{\tabcolsep}{3pt}
\begin{tabular*}{\textwidth}{@{\extracolsep{\fill}}lrrrrrr@{}}
\toprule
\textbf{Learner} & \multicolumn{2}{c}{\shortstack{\textbf{CBIS-DDSM}\\(AUC)}} & \multicolumn{2}{c}{\shortstack{\textbf{CheXpert}\\(AUC)}} & \multicolumn{2}{c}{\shortstack{\textbf{CSGO Skin Price}\\(AUC)}} \\
\cmidrule(lr){2-3}\cmidrule(lr){4-5}\cmidrule(lr){6-7}
 & \textbf{Frozen} & \textbf{LightTAR} & \textbf{Frozen} & \textbf{LightTAR} & \textbf{Frozen} & \textbf{LightTAR} \\
\midrule
CatBoost & $0.849\,\pm\,0.019$ & $0.863\,\pm\,0.025$ & $0.772\,\pm\,0.003$ & $0.777\,\pm\,0.006$ & $0.865\,\pm\,0.017$ & $0.875\,\pm\,0.018$ \\
LightGBM & $0.823\,\pm\,0.022$ & $0.850\,\pm\,0.024$ & $0.767\,\pm\,0.005$ & $0.774\,\pm\,0.005$ & $0.832\,\pm\,0.028$ & $0.832\,\pm\,0.029$ \\
XGBoost & $0.810\,\pm\,0.024$ & $0.834\,\pm\,0.031$ & $0.722\,\pm\,0.009$ & $0.737\,\pm\,0.006$ & $0.843\,\pm\,0.011$ & $0.840\,\pm\,0.031$ \\
Random Forest & $0.832\,\pm\,0.021$ & $0.850\,\pm\,0.023$ & $0.711\,\pm\,0.007$ & $0.726\,\pm\,0.005$ & $0.841\,\pm\,0.012$ & $0.853\,\pm\,0.018$ \\
TabFM & $0.900\,\pm\,0.008$ & $0.914\,\pm\,0.015$ & $0.809\,\pm\,0.006$ & $0.817\,\pm\,0.006$ & $0.920\,\pm\,0.008$ & $0.925\,\pm\,0.011$ \\
RealMLP & $0.845\,\pm\,0.012$ & $0.864\,\pm\,0.016$ & $0.741\,\pm\,0.006$ & $0.757\,\pm\,0.006$ & $0.859\,\pm\,0.019$ & $0.866\,\pm\,0.018$ \\
TabM & $0.868\,\pm\,0.018$ & $0.882\,\pm\,0.017$ & $0.745\,\pm\,0.007$ & $0.758\,\pm\,0.006$ & $0.864\,\pm\,0.004$ & $0.857\,\pm\,0.025$ \\
TabICLv2 & $0.898\,\pm\,0.006$ & $0.900\,\pm\,0.018$ & $0.789\,\pm\,0.007$ & $0.796\,\pm\,0.008$ & $0.910\,\pm\,0.007$ & $0.912\,\pm\,0.011$ \\
TabDPT 1.1 & $0.882\,\pm\,0.007$ & $0.896\,\pm\,0.016$ & $0.778\,\pm\,0.003$ & $0.783\,\pm\,0.003$ & $0.895\,\pm\,0.012$ & $0.894\,\pm\,0.010$ \\
TabDPT 1.3 & $0.886\,\pm\,0.010$ & $0.899\,\pm\,0.016$ & $0.788\,\pm\,0.007$ & $0.795\,\pm\,0.008$ & $0.911\,\pm\,0.007$ & $0.913\,\pm\,0.011$ \\
TabPFN v2 & $0.873\,\pm\,0.013$ & $0.885\,\pm\,0.023$ & $0.784\,\pm\,0.008$ & $0.790\,\pm\,0.005$ & $0.904\,\pm\,0.009$ & $0.901\,\pm\,0.007$ \\
TabPFN 2.5 & $0.883\,\pm\,0.012$ & $0.893\,\pm\,0.021$ & $0.786\,\pm\,0.008$ & $0.796\,\pm\,0.008$ & $0.908\,\pm\,0.009$ & $0.910\,\pm\,0.006$ \\
TabPFN v3 & $0.892\,\pm\,0.013$ & $0.898\,\pm\,0.022$ & $0.789\,\pm\,0.006$ & $0.795\,\pm\,0.005$ & $0.909\,\pm\,0.007$ & $0.910\,\pm\,0.007$ \\
TabPFN 3.5 & $0.899\,\pm\,0.010$ & $0.907\,\pm\,0.019$ & $0.794\,\pm\,0.007$ & $0.800\,\pm\,0.006$ & $0.922\,\pm\,0.005$ & $0.919\,\pm\,0.011$ \\
TabPFN 3.5 Fast & $0.902\,\pm\,0.011$ & $0.906\,\pm\,0.017$ & $0.789\,\pm\,0.006$ & $0.795\,\pm\,0.006$ & $0.914\,\pm\,0.005$ & $0.913\,\pm\,0.008$ \\
Kumo Small & $0.898\,\pm\,0.009$ & $0.909\,\pm\,0.016$ & $0.786\,\pm\,0.006$ & $0.793\,\pm\,0.006$ & $0.916\,\pm\,0.010$ & $0.918\,\pm\,0.009$ \\
Kumo Medium & $0.898\,\pm\,0.010$ & $0.908\,\pm\,0.015$ & $0.784\,\pm\,0.005$ & $0.793\,\pm\,0.005$ & $0.918\,\pm\,0.004$ & $0.920\,\pm\,0.011$ \\
Kumo Large & $0.903\,\pm\,0.008$ & $0.914\,\pm\,0.014$ & $0.790\,\pm\,0.005$ & $0.800\,\pm\,0.007$ & $0.919\,\pm\,0.004$ & $0.921\,\pm\,0.010$ \\
Causilo & $0.895\,\pm\,0.007$ & $0.905\,\pm\,0.015$ & $0.791\,\pm\,0.008$ & $0.797\,\pm\,0.008$ & $0.914\,\pm\,0.006$ & $0.913\,\pm\,0.010$ \\
TabLDM & $0.896\,\pm\,0.008$ & $0.902\,\pm\,0.019$ & $0.789\,\pm\,0.008$ & $0.796\,\pm\,0.008$ & $0.908\,\pm\,0.006$ & $0.908\,\pm\,0.012$ \\
EXAONE & $0.893\,\pm\,0.008$ & $0.904\,\pm\,0.014$ & $0.785\,\pm\,0.006$ & $0.793\,\pm\,0.006$ & $0.908\,\pm\,0.010$ & $0.908\,\pm\,0.009$ \\
\bottomrule
\end{tabular*}
\par\medskip
\centering
\caption{\textbf{MulTaBench image (3/7).}}
\label{tab:mtb_raw_image_3}
\small
\setlength{\tabcolsep}{3pt}
\begin{tabular*}{\textwidth}{@{\extracolsep{\fill}}lrrrrrr@{}}
\toprule
\textbf{Learner} & \multicolumn{2}{c}{\shortstack{\textbf{Flower Bouquets}\\(AUC)}} & \multicolumn{2}{c}{\shortstack{\textbf{Glaucoma SMDG}\\(AUC)}} & \multicolumn{2}{c}{\shortstack{\textbf{HuBMAP HPA}\\(AUC)}} \\
\cmidrule(lr){2-3}\cmidrule(lr){4-5}\cmidrule(lr){6-7}
 & \textbf{Frozen} & \textbf{LightTAR} & \textbf{Frozen} & \textbf{LightTAR} & \textbf{Frozen} & \textbf{LightTAR} \\
\midrule
CatBoost & $0.612\,\pm\,0.026$ & $0.623\,\pm\,0.055$ & $0.920\,\pm\,0.015$ & $0.931\,\pm\,0.010$ & $0.957\,\pm\,0.002$ & $0.965\,\pm\,0.002$ \\
LightGBM & $0.601\,\pm\,0.050$ & $0.622\,\pm\,0.075$ & $0.921\,\pm\,0.018$ & $0.929\,\pm\,0.013$ & $0.947\,\pm\,0.002$ & $0.957\,\pm\,0.002$ \\
XGBoost & $0.571\,\pm\,0.040$ & $0.606\,\pm\,0.049$ & $0.905\,\pm\,0.023$ & $0.926\,\pm\,0.019$ & $0.947\,\pm\,0.003$ & $0.956\,\pm\,0.003$ \\
Random Forest & $0.620\,\pm\,0.049$ & $0.637\,\pm\,0.039$ & $0.898\,\pm\,0.021$ & $0.910\,\pm\,0.009$ & $0.936\,\pm\,0.005$ & $0.944\,\pm\,0.004$ \\
TabFM & $0.684\,\pm\,0.021$ & $0.661\,\pm\,0.026$ & $0.949\,\pm\,0.009$ & $0.962\,\pm\,0.008$ & $0.989\,\pm\,0.001$ & $0.995\,\pm\,0.001$ \\
RealMLP & $0.603\,\pm\,0.066$ & $0.623\,\pm\,0.070$ & $0.919\,\pm\,0.019$ & $0.933\,\pm\,0.016$ & $0.966\,\pm\,0.002$ & $0.975\,\pm\,0.001$ \\
TabM & $0.621\,\pm\,0.046$ & $0.619\,\pm\,0.038$ & $0.925\,\pm\,0.018$ & $0.938\,\pm\,0.014$ & $0.970\,\pm\,0.002$ & $0.977\,\pm\,0.001$ \\
TabICLv2 & $0.680\,\pm\,0.036$ & $0.663\,\pm\,0.020$ & $0.945\,\pm\,0.010$ & $0.955\,\pm\,0.007$ & $0.982\,\pm\,0.001$ & $0.991\,\pm\,0.001$ \\
TabDPT 1.1 & $0.690\,\pm\,0.054$ & $0.687\,\pm\,0.022$ & $0.938\,\pm\,0.012$ & $0.943\,\pm\,0.011$ & $0.961\,\pm\,0.002$ & $0.966\,\pm\,0.003$ \\
TabDPT 1.3 & $0.688\,\pm\,0.035$ & $0.687\,\pm\,0.013$ & $0.942\,\pm\,0.012$ & $0.952\,\pm\,0.008$ & $0.977\,\pm\,0.001$ & $0.985\,\pm\,0.000$ \\
TabPFN v2 & $0.658\,\pm\,0.018$ & $0.653\,\pm\,0.034$ & $0.935\,\pm\,0.014$ & $0.942\,\pm\,0.011$ & $0.958\,\pm\,0.002$ & $0.968\,\pm\,0.001$ \\
TabPFN 2.5 & $0.662\,\pm\,0.029$ & $0.661\,\pm\,0.028$ & $0.939\,\pm\,0.011$ & $0.950\,\pm\,0.009$ & $0.969\,\pm\,0.002$ & $0.979\,\pm\,0.001$ \\
TabPFN v3 & $0.667\,\pm\,0.021$ & $0.667\,\pm\,0.028$ & $0.944\,\pm\,0.009$ & $0.953\,\pm\,0.006$ & $0.981\,\pm\,0.001$ & $0.989\,\pm\,0.001$ \\
TabPFN 3.5 & $0.687\,\pm\,0.023$ & $0.666\,\pm\,0.022$ & $0.947\,\pm\,0.009$ & $0.957\,\pm\,0.006$ & $0.986\,\pm\,0.001$ & $0.993\,\pm\,0.001$ \\
TabPFN 3.5 Fast & $0.683\,\pm\,0.026$ & $0.663\,\pm\,0.021$ & $0.945\,\pm\,0.008$ & $0.955\,\pm\,0.005$ & $0.985\,\pm\,0.001$ & $0.992\,\pm\,0.001$ \\
Kumo Small & $0.691\,\pm\,0.031$ & $0.669\,\pm\,0.016$ & $0.945\,\pm\,0.009$ & $0.954\,\pm\,0.007$ & $0.984\,\pm\,0.001$ & $0.991\,\pm\,0.001$ \\
Kumo Medium & $0.697\,\pm\,0.024$ & $0.671\,\pm\,0.017$ & $0.948\,\pm\,0.009$ & $0.956\,\pm\,0.007$ & $0.986\,\pm\,0.001$ & $0.993\,\pm\,0.001$ \\
Kumo Large & $0.680\,\pm\,0.028$ & $0.662\,\pm\,0.020$ & $0.949\,\pm\,0.008$ & $0.958\,\pm\,0.006$ & $0.988\,\pm\,0.001$ & $0.994\,\pm\,0.001$ \\
Causilo & $0.686\,\pm\,0.034$ & $0.663\,\pm\,0.034$ & $0.946\,\pm\,0.009$ & $0.954\,\pm\,0.007$ & $0.983\,\pm\,0.001$ & $0.991\,\pm\,0.001$ \\
TabLDM & $0.681\,\pm\,0.042$ & $0.657\,\pm\,0.026$ & $0.943\,\pm\,0.010$ & $0.952\,\pm\,0.007$ & $0.980\,\pm\,0.001$ & $0.988\,\pm\,0.001$ \\
EXAONE & $0.687\,\pm\,0.034$ & $0.677\,\pm\,0.023$ & $0.940\,\pm\,0.011$ & $0.949\,\pm\,0.008$ & $0.977\,\pm\,0.001$ & $0.986\,\pm\,0.001$ \\
\bottomrule
\end{tabular*}
\end{table*}

\begin{table*}[!tbp]
\centering
\caption{\textbf{MulTaBench image (4/7).}}
\label{tab:mtb_raw_image_4}
\small
\setlength{\tabcolsep}{3pt}
\begin{tabular*}{\textwidth}{@{\extracolsep{\fill}}lrrrrrr@{}}
\toprule
\textbf{Learner} & \multicolumn{2}{c}{\shortstack{\textbf{Justin Instagram}\\(AUC)}} & \multicolumn{2}{c}{\shortstack{\textbf{PetFinder}\\(AUC)}} & \multicolumn{2}{c}{\shortstack{\textbf{ZooScan Zooplankton}\\(AUC)}} \\
\cmidrule(lr){2-3}\cmidrule(lr){4-5}\cmidrule(lr){6-7}
 & \textbf{Frozen} & \textbf{LightTAR} & \textbf{Frozen} & \textbf{LightTAR} & \textbf{Frozen} & \textbf{LightTAR} \\
\midrule
CatBoost & $0.950\,\pm\,0.007$ & $0.957\,\pm\,0.010$ & $0.831\,\pm\,0.005$ & $0.835\,\pm\,0.006$ & $0.983\,\pm\,0.002$ & $0.984\,\pm\,0.002$ \\
LightGBM & $0.949\,\pm\,0.009$ & $0.953\,\pm\,0.006$ & $0.812\,\pm\,0.004$ & $0.823\,\pm\,0.004$ & $0.978\,\pm\,0.003$ & $0.979\,\pm\,0.003$ \\
XGBoost & $0.946\,\pm\,0.007$ & $0.946\,\pm\,0.007$ & $0.802\,\pm\,0.006$ & $0.812\,\pm\,0.007$ & $0.977\,\pm\,0.002$ & $0.978\,\pm\,0.002$ \\
Random Forest & $0.922\,\pm\,0.014$ & $0.934\,\pm\,0.011$ & $0.799\,\pm\,0.007$ & $0.815\,\pm\,0.005$ & $0.969\,\pm\,0.002$ & $0.971\,\pm\,0.002$ \\
TabFM & $0.971\,\pm\,0.007$ & $0.975\,\pm\,0.007$ & $0.876\,\pm\,0.004$ & $0.879\,\pm\,0.005$ & $0.993\,\pm\,0.001$ & $0.993\,\pm\,0.001$ \\
RealMLP & $0.948\,\pm\,0.015$ & $0.951\,\pm\,0.014$ & $0.835\,\pm\,0.006$ & $0.834\,\pm\,0.008$ & $0.985\,\pm\,0.003$ & $0.984\,\pm\,0.003$ \\
TabM & $0.957\,\pm\,0.017$ & $0.960\,\pm\,0.011$ & $0.845\,\pm\,0.004$ & $0.848\,\pm\,0.003$ & $0.986\,\pm\,0.002$ & $0.986\,\pm\,0.002$ \\
TabICLv2 & $0.967\,\pm\,0.007$ & $0.971\,\pm\,0.006$ & $0.864\,\pm\,0.006$ & $0.863\,\pm\,0.007$ & $0.990\,\pm\,0.001$ & $0.990\,\pm\,0.002$ \\
TabDPT 1.1 & $0.955\,\pm\,0.008$ & $0.962\,\pm\,0.011$ & $0.828\,\pm\,0.007$ & $0.840\,\pm\,0.004$ & $0.987\,\pm\,0.002$ & $0.987\,\pm\,0.001$ \\
TabDPT 1.3 & $0.964\,\pm\,0.009$ & $0.967\,\pm\,0.009$ & $0.843\,\pm\,0.006$ & $0.851\,\pm\,0.005$ & $0.989\,\pm\,0.002$ & $0.989\,\pm\,0.002$ \\
TabPFN v2 & $0.950\,\pm\,0.009$ & $0.950\,\pm\,0.009$ & $0.840\,\pm\,0.005$ & $0.843\,\pm\,0.004$ & $0.984\,\pm\,0.002$ & $0.984\,\pm\,0.001$ \\
TabPFN 2.5 & $0.959\,\pm\,0.008$ & $0.963\,\pm\,0.008$ & $0.849\,\pm\,0.004$ & $0.853\,\pm\,0.005$ & $0.988\,\pm\,0.002$ & $0.988\,\pm\,0.002$ \\
TabPFN v3 & $0.965\,\pm\,0.009$ & $0.969\,\pm\,0.009$ & $0.846\,\pm\,0.005$ & $0.851\,\pm\,0.004$ & $0.989\,\pm\,0.001$ & $0.989\,\pm\,0.002$ \\
TabPFN 3.5 & $0.969\,\pm\,0.009$ & $0.974\,\pm\,0.009$ & $0.867\,\pm\,0.007$ & $0.873\,\pm\,0.006$ & $0.992\,\pm\,0.001$ & $0.992\,\pm\,0.001$ \\
TabPFN 3.5 Fast & $0.969\,\pm\,0.010$ & $0.972\,\pm\,0.010$ & $0.864\,\pm\,0.007$ & $0.866\,\pm\,0.007$ & $0.991\,\pm\,0.001$ & $0.991\,\pm\,0.001$ \\
Kumo Small & $0.968\,\pm\,0.008$ & $0.971\,\pm\,0.006$ & $0.864\,\pm\,0.005$ & $0.867\,\pm\,0.006$ & $0.991\,\pm\,0.001$ & $0.991\,\pm\,0.001$ \\
Kumo Medium & $0.970\,\pm\,0.009$ & $0.973\,\pm\,0.008$ & $0.865\,\pm\,0.005$ & $0.875\,\pm\,0.005$ & $0.992\,\pm\,0.001$ & $0.992\,\pm\,0.001$ \\
Kumo Large & $0.970\,\pm\,0.009$ & $0.974\,\pm\,0.008$ & $0.869\,\pm\,0.005$ & $0.877\,\pm\,0.005$ & $0.992\,\pm\,0.001$ & $0.992\,\pm\,0.001$ \\
Causilo & $0.968\,\pm\,0.008$ & $0.972\,\pm\,0.007$ & $0.869\,\pm\,0.005$ & $0.872\,\pm\,0.006$ & $0.991\,\pm\,0.001$ & $0.991\,\pm\,0.001$ \\
TabLDM & $0.965\,\pm\,0.008$ & $0.969\,\pm\,0.008$ & $0.861\,\pm\,0.006$ & $0.860\,\pm\,0.006$ & $0.989\,\pm\,0.001$ & $0.990\,\pm\,0.002$ \\
EXAONE & $0.965\,\pm\,0.011$ & $0.969\,\pm\,0.009$ & $0.860\,\pm\,0.006$ & $0.864\,\pm\,0.006$ & $0.990\,\pm\,0.001$ & $0.990\,\pm\,0.001$ \\
\bottomrule
\end{tabular*}
\par\medskip
\centering
\caption{\textbf{MulTaBench image (5/7).}}
\label{tab:mtb_raw_image_5}
\small
\setlength{\tabcolsep}{3pt}
\begin{tabular*}{\textwidth}{@{\extracolsep{\fill}}lrrrrrr@{}}
\toprule
\textbf{Learner} & \multicolumn{2}{c}{\shortstack{\textbf{Amazon Best Seller}\\($R^2$)}} & \multicolumn{2}{c}{\shortstack{\textbf{Amazon Packages}\\($R^2$)}} & \multicolumn{2}{c}{\shortstack{\textbf{H\&M Fashion}\\($R^2$)}} \\
\cmidrule(lr){2-3}\cmidrule(lr){4-5}\cmidrule(lr){6-7}
 & \textbf{Frozen} & \textbf{LightTAR} & \textbf{Frozen} & \textbf{LightTAR} & \textbf{Frozen} & \textbf{LightTAR} \\
\midrule
CatBoost & $0.526\,\pm\,0.045$ & $0.530\,\pm\,0.060$ & $0.495\,\pm\,0.049$ & $0.559\,\pm\,0.038$ & $0.274\,\pm\,0.013$ & $0.275\,\pm\,0.015$ \\
LightGBM & $0.486\,\pm\,0.049$ & $0.504\,\pm\,0.053$ & $0.474\,\pm\,0.043$ & $0.540\,\pm\,0.040$ & $0.257\,\pm\,0.015$ & $0.263\,\pm\,0.012$ \\
XGBoost & $0.427\,\pm\,0.067$ & $0.447\,\pm\,0.065$ & $0.414\,\pm\,0.042$ & $0.504\,\pm\,0.044$ & $0.222\,\pm\,0.016$ & $0.228\,\pm\,0.014$ \\
Random Forest & $0.473\,\pm\,0.038$ & $0.466\,\pm\,0.051$ & $0.403\,\pm\,0.052$ & $0.518\,\pm\,0.041$ & $0.251\,\pm\,0.013$ & $0.262\,\pm\,0.014$ \\
TabFM & $0.637\,\pm\,0.044$ & $0.640\,\pm\,0.042$ & $0.556\,\pm\,0.045$ & $0.587\,\pm\,0.036$ & $0.311\,\pm\,0.015$ & $0.308\,\pm\,0.013$ \\
RealMLP & $0.553\,\pm\,0.059$ & $0.523\,\pm\,0.079$ & $0.486\,\pm\,0.035$ & $0.526\,\pm\,0.024$ & $0.232\,\pm\,0.007$ & $0.244\,\pm\,0.016$ \\
TabM & $0.567\,\pm\,0.053$ & $0.578\,\pm\,0.059$ & $0.530\,\pm\,0.051$ & $0.569\,\pm\,0.039$ & $0.284\,\pm\,0.013$ & $0.281\,\pm\,0.018$ \\
TabICLv2 & $0.617\,\pm\,0.043$ & $0.630\,\pm\,0.050$ & $0.548\,\pm\,0.040$ & $0.589\,\pm\,0.040$ & $0.306\,\pm\,0.015$ & $0.308\,\pm\,0.015$ \\
TabDPT 1.1 & $0.625\,\pm\,0.048$ & $0.629\,\pm\,0.047$ & $0.537\,\pm\,0.040$ & $0.581\,\pm\,0.040$ & $0.289\,\pm\,0.014$ & $0.291\,\pm\,0.015$ \\
TabDPT 1.3 & $0.632\,\pm\,0.037$ & $0.633\,\pm\,0.045$ & $0.557\,\pm\,0.043$ & $0.592\,\pm\,0.036$ & $0.290\,\pm\,0.012$ & $0.294\,\pm\,0.012$ \\
TabPFN v2 & $0.601\,\pm\,0.038$ & $0.582\,\pm\,0.054$ & $0.528\,\pm\,0.042$ & $0.570\,\pm\,0.038$ & $0.274\,\pm\,0.011$ & $0.276\,\pm\,0.014$ \\
TabPFN 2.5 & $0.612\,\pm\,0.052$ & $0.601\,\pm\,0.050$ & $0.541\,\pm\,0.045$ & $0.581\,\pm\,0.040$ & $0.296\,\pm\,0.011$ & $0.296\,\pm\,0.013$ \\
TabPFN v3 & $0.625\,\pm\,0.063$ & $0.630\,\pm\,0.051$ & $0.550\,\pm\,0.043$ & $0.583\,\pm\,0.038$ & $0.295\,\pm\,0.017$ & $0.299\,\pm\,0.020$ \\
TabPFN 3.5 & $0.656\,\pm\,0.048$ & $0.657\,\pm\,0.048$ & $0.561\,\pm\,0.045$ & $0.593\,\pm\,0.038$ & $0.319\,\pm\,0.016$ & $0.321\,\pm\,0.015$ \\
TabPFN 3.5 Fast & $0.647\,\pm\,0.050$ & $0.646\,\pm\,0.048$ & $0.557\,\pm\,0.041$ & $0.590\,\pm\,0.037$ & $0.316\,\pm\,0.014$ & $0.316\,\pm\,0.016$ \\
Kumo Small & $0.648\,\pm\,0.051$ & $0.652\,\pm\,0.048$ & $0.556\,\pm\,0.045$ & $0.590\,\pm\,0.037$ & $0.321\,\pm\,0.014$ & $0.318\,\pm\,0.016$ \\
Kumo Medium & $0.653\,\pm\,0.054$ & $0.660\,\pm\,0.048$ & $0.559\,\pm\,0.046$ & $0.594\,\pm\,0.038$ & $0.323\,\pm\,0.014$ & $0.323\,\pm\,0.018$ \\
Kumo Large & $0.662\,\pm\,0.056$ & $0.665\,\pm\,0.050$ & $0.565\,\pm\,0.049$ & $0.597\,\pm\,0.040$ & $0.325\,\pm\,0.015$ & $0.325\,\pm\,0.017$ \\
Causilo & $0.652\,\pm\,0.052$ & $0.652\,\pm\,0.054$ & $0.562\,\pm\,0.043$ & $0.594\,\pm\,0.036$ & $0.313\,\pm\,0.015$ & $0.315\,\pm\,0.017$ \\
TabLDM & $0.636\,\pm\,0.047$ & $0.632\,\pm\,0.050$ & $0.552\,\pm\,0.043$ & $0.590\,\pm\,0.038$ & $0.312\,\pm\,0.012$ & $0.314\,\pm\,0.017$ \\
EXAONE & $0.626\,\pm\,0.051$ & $0.616\,\pm\,0.057$ & $0.546\,\pm\,0.042$ & $0.584\,\pm\,0.038$ & $0.312\,\pm\,0.011$ & $0.311\,\pm\,0.014$ \\
\bottomrule
\end{tabular*}
\end{table*}

\begin{table*}[!tbp]
\centering
\caption{\textbf{MulTaBench image (6/7).}}
\label{tab:mtb_raw_image_6}
\small
\setlength{\tabcolsep}{3pt}
\begin{tabular*}{\textwidth}{@{\extracolsep{\fill}}lrrrrrr@{}}
\toprule
\textbf{Learner} & \multicolumn{2}{c}{\shortstack{\textbf{Khaadi Clothes}\\($R^2$)}} & \multicolumn{2}{c}{\shortstack{\textbf{Letterboxd Movies}\\($R^2$)}} & \multicolumn{2}{c}{\shortstack{\textbf{Mango Mass}\\($R^2$)}} \\
\cmidrule(lr){2-3}\cmidrule(lr){4-5}\cmidrule(lr){6-7}
 & \textbf{Frozen} & \textbf{LightTAR} & \textbf{Frozen} & \textbf{LightTAR} & \textbf{Frozen} & \textbf{LightTAR} \\
\midrule
CatBoost & $0.419\,\pm\,0.378$ & $0.387\,\pm\,0.644$ & $0.449\,\pm\,0.025$ & $0.448\,\pm\,0.029$ & $0.505\,\pm\,0.052$ & $0.569\,\pm\,0.014$ \\
LightGBM & $0.161\,\pm\,0.759$ & $0.129\,\pm\,1.203$ & $0.445\,\pm\,0.029$ & $0.432\,\pm\,0.032$ & $0.478\,\pm\,0.058$ & $0.565\,\pm\,0.050$ \\
XGBoost & $0.325\,\pm\,0.172$ & $0.139\,\pm\,0.907$ & $0.407\,\pm\,0.027$ & $0.390\,\pm\,0.038$ & $0.358\,\pm\,0.129$ & $0.489\,\pm\,0.056$ \\
Random Forest & $0.312\,\pm\,0.405$ & $0.190\,\pm\,0.946$ & $0.412\,\pm\,0.028$ & $0.403\,\pm\,0.030$ & $0.421\,\pm\,0.079$ & $0.527\,\pm\,0.039$ \\
TabFM & $0.866\,\pm\,0.087$ & $0.840\,\pm\,0.089$ & $0.498\,\pm\,0.038$ & $0.501\,\pm\,0.031$ & $0.640\,\pm\,0.034$ & $0.686\,\pm\,0.037$ \\
RealMLP & $0.734\,\pm\,0.164$ & $0.557\,\pm\,0.324$ & $0.418\,\pm\,0.043$ & $0.426\,\pm\,0.017$ & $0.533\,\pm\,0.085$ & $0.552\,\pm\,0.095$ \\
TabM & $0.517\,\pm\,0.571$ & $0.588\,\pm\,0.425$ & $0.468\,\pm\,0.031$ & $0.464\,\pm\,0.025$ & $0.559\,\pm\,0.058$ & $0.567\,\pm\,0.062$ \\
TabICLv2 & $0.738\,\pm\,0.345$ & $0.880\,\pm\,0.084$ & $0.468\,\pm\,0.029$ & $0.478\,\pm\,0.030$ & $0.620\,\pm\,0.024$ & $0.680\,\pm\,0.032$ \\
TabDPT 1.1 & $0.663\,\pm\,0.093$ & $0.672\,\pm\,0.108$ & $0.389\,\pm\,0.030$ & $0.389\,\pm\,0.023$ & $0.624\,\pm\,0.026$ & $0.640\,\pm\,0.022$ \\
TabDPT 1.3 & $0.769\,\pm\,0.198$ & $0.835\,\pm\,0.092$ & $0.438\,\pm\,0.028$ & $0.437\,\pm\,0.029$ & $0.617\,\pm\,0.026$ & $0.648\,\pm\,0.032$ \\
TabPFN v2 & $0.661\,\pm\,0.208$ & $0.608\,\pm\,0.314$ & $0.466\,\pm\,0.033$ & $0.469\,\pm\,0.036$ & $0.580\,\pm\,0.035$ & $0.654\,\pm\,0.044$ \\
TabPFN 2.5 & $0.308\,\pm\,0.536$ & $0.378\,\pm\,0.476$ & $0.483\,\pm\,0.030$ & $0.486\,\pm\,0.032$ & $0.600\,\pm\,0.031$ & $0.650\,\pm\,0.026$ \\
TabPFN v3 & $0.792\,\pm\,0.167$ & $0.808\,\pm\,0.145$ & $0.484\,\pm\,0.029$ & $0.489\,\pm\,0.031$ & $0.620\,\pm\,0.042$ & $0.666\,\pm\,0.031$ \\
TabPFN 3.5 & $0.884\,\pm\,0.060$ & $0.888\,\pm\,0.056$ & $0.491\,\pm\,0.034$ & $0.498\,\pm\,0.035$ & $0.643\,\pm\,0.032$ & $0.703\,\pm\,0.039$ \\
TabPFN 3.5 Fast & $0.891\,\pm\,0.060$ & $0.896\,\pm\,0.039$ & $0.484\,\pm\,0.032$ & $0.494\,\pm\,0.032$ & $0.631\,\pm\,0.037$ & $0.678\,\pm\,0.020$ \\
Kumo Small & $0.750\,\pm\,0.248$ & $0.776\,\pm\,0.196$ & $0.499\,\pm\,0.032$ & $0.501\,\pm\,0.034$ & $0.646\,\pm\,0.043$ & $0.691\,\pm\,0.029$ \\
Kumo Medium & $0.882\,\pm\,0.061$ & $0.861\,\pm\,0.083$ & $0.507\,\pm\,0.033$ & $0.510\,\pm\,0.032$ & $0.647\,\pm\,0.037$ & $0.692\,\pm\,0.039$ \\
Kumo Large & $0.878\,\pm\,0.065$ & $0.881\,\pm\,0.065$ & $0.509\,\pm\,0.031$ & $0.514\,\pm\,0.033$ & $0.660\,\pm\,0.042$ & $0.697\,\pm\,0.036$ \\
Causilo & $0.887\,\pm\,0.081$ & $0.876\,\pm\,0.095$ & $0.494\,\pm\,0.032$ & $0.497\,\pm\,0.034$ & $0.638\,\pm\,0.040$ & $0.691\,\pm\,0.030$ \\
TabLDM & $0.847\,\pm\,0.148$ & $0.883\,\pm\,0.079$ & $0.480\,\pm\,0.035$ & $0.475\,\pm\,0.037$ & $0.618\,\pm\,0.034$ & $0.683\,\pm\,0.039$ \\
EXAONE & $0.733\,\pm\,0.380$ & $0.786\,\pm\,0.244$ & $0.500\,\pm\,0.030$ & $0.499\,\pm\,0.030$ & $0.635\,\pm\,0.039$ & $0.673\,\pm\,0.018$ \\
\bottomrule
\end{tabular*}
\par\medskip
\centering
\caption{\textbf{MulTaBench image (7/7).}}
\label{tab:mtb_raw_image_7}
\small
\setlength{\tabcolsep}{3pt}
\begin{tabular*}{\textwidth}{@{\extracolsep{\fill}}lrrrr@{}}
\toprule
\textbf{Learner} & \multicolumn{2}{c}{\shortstack{\textbf{MKPhoto Bots}\\($R^2$)}} & \multicolumn{2}{c}{\shortstack{\textbf{Painting Price}\\($R^2$)}} \\
\cmidrule(lr){2-3}\cmidrule(lr){4-5}
 & \textbf{Frozen} & \textbf{LightTAR} & \textbf{Frozen} & \textbf{LightTAR} \\
\midrule
CatBoost & $0.338\,\pm\,0.029$ & $0.355\,\pm\,0.025$ & $0.288\,\pm\,0.077$ & $0.321\,\pm\,0.074$ \\
LightGBM & $0.314\,\pm\,0.022$ & $0.335\,\pm\,0.020$ & $0.243\,\pm\,0.044$ & $0.285\,\pm\,0.073$ \\
XGBoost & $0.280\,\pm\,0.038$ & $0.313\,\pm\,0.025$ & $0.186\,\pm\,0.083$ & $0.256\,\pm\,0.053$ \\
Random Forest & $0.345\,\pm\,0.037$ & $0.366\,\pm\,0.023$ & $0.239\,\pm\,0.050$ & $0.243\,\pm\,0.069$ \\
TabFM & $0.512\,\pm\,0.032$ & $0.520\,\pm\,0.027$ & $0.446\,\pm\,0.062$ & $0.475\,\pm\,0.063$ \\
RealMLP & $0.309\,\pm\,0.060$ & $0.301\,\pm\,0.040$ & $0.159\,\pm\,0.089$ & $0.228\,\pm\,0.045$ \\
TabM & $0.349\,\pm\,0.048$ & $0.368\,\pm\,0.039$ & $0.340\,\pm\,0.086$ & $0.363\,\pm\,0.082$ \\
TabICLv2 & $0.428\,\pm\,0.037$ & $0.431\,\pm\,0.034$ & $0.398\,\pm\,0.066$ & $0.399\,\pm\,0.076$ \\
TabDPT 1.1 & $0.428\,\pm\,0.024$ & $0.447\,\pm\,0.032$ & $0.331\,\pm\,0.076$ & $0.345\,\pm\,0.075$ \\
TabDPT 1.3 & $0.434\,\pm\,0.035$ & $0.446\,\pm\,0.027$ & $0.365\,\pm\,0.063$ & $0.390\,\pm\,0.055$ \\
TabPFN v2 & $0.313\,\pm\,0.030$ & $0.339\,\pm\,0.023$ & $0.292\,\pm\,0.087$ & $0.314\,\pm\,0.080$ \\
TabPFN 2.5 & $0.369\,\pm\,0.024$ & $0.374\,\pm\,0.024$ & $0.371\,\pm\,0.065$ & $0.379\,\pm\,0.076$ \\
TabPFN v3 & $0.432\,\pm\,0.043$ & $0.441\,\pm\,0.031$ & $0.389\,\pm\,0.092$ & $0.411\,\pm\,0.077$ \\
TabPFN 3.5 & $0.491\,\pm\,0.032$ & $0.505\,\pm\,0.032$ & $0.426\,\pm\,0.194$ & $0.401\,\pm\,0.217$ \\
TabPFN 3.5 Fast & $0.474\,\pm\,0.035$ & $0.471\,\pm\,0.031$ & $0.432\,\pm\,0.160$ & $0.424\,\pm\,0.209$ \\
Kumo Small & $0.469\,\pm\,0.030$ & $0.478\,\pm\,0.024$ & $0.415\,\pm\,0.106$ & $0.452\,\pm\,0.073$ \\
Kumo Medium & $0.503\,\pm\,0.032$ & $0.507\,\pm\,0.026$ & $0.436\,\pm\,0.100$ & $0.469\,\pm\,0.068$ \\
Kumo Large & $0.524\,\pm\,0.032$ & $0.528\,\pm\,0.028$ & $0.455\,\pm\,0.077$ & $0.491\,\pm\,0.063$ \\
Causilo & $0.471\,\pm\,0.029$ & $0.481\,\pm\,0.023$ & $0.392\,\pm\,0.107$ & $0.417\,\pm\,0.099$ \\
TabLDM & $0.432\,\pm\,0.030$ & $0.444\,\pm\,0.033$ & $0.376\,\pm\,0.098$ & $0.392\,\pm\,0.085$ \\
EXAONE & $0.422\,\pm\,0.030$ & $0.427\,\pm\,0.028$ & $0.402\,\pm\,0.100$ & $0.421\,\pm\,0.095$ \\
\bottomrule
\end{tabular*}
\end{table*}

\section{Extended Ablation Results}
\label{app:ablations}

\paragraph{Elo gains in Table~\ref{tab:mechanism_ablations}.}
We compare the eight representations using the five learners in
Section~\ref{sec:analysis}. All comparisons use the same five splits
for each supported dataset (98 learner--dataset pairs on text and 100 on
image). Within each benchmark, we fit separate Bradley--Terry ratings
for classification and regression. Each fit includes all 40
learner--representation pairs and uses the split-level comparisons
described in Appendix~\ref{app:elo}.

Let $\bar E_v$ be the mean Elo rating of representation $v$ across the
five learners. The percentage reported for $v$ is
\[
100\,\frac{\bar E_v-\bar E_{\mathrm{Frozen}}}
              {\bar E_{\mathrm{LightTAR}}-\bar E_{\mathrm{Frozen}}}.
\]

\paragraph{Projection variants.}
The shuffled-label control permutes the labels used to fit the projection.
It keeps the true labels for downstream training and all other LightTAR
settings fixed. Joint PCA fits 35 components to the concatenated layers.
The final-layer control fits LightTAR's five supervised directions to
the final layer alone and retains the 30-component anchor. Removing the
anchor retains the two-layer supervised projection but reduces the
feature count from 35 to five.

Anchored textbook PLS adds five final-layer NIPALS components to the
PCA anchor. Without an anchor, textbook PLS uses both layers and selects
the number of components from $\{5,10,15,20,30,35\}$ on an inner
80/20 training--validation split.

\paragraph{Image and text contributions.}
Nine MulTaBench image datasets contain both image and text columns
(four classification and five regression datasets). Across the same
10 learners and five splits, applying LightTAR to images while keeping
text at its Frozen representation improves mean AUC by 0.72 points
and mean $R^2$ by 1.08 points over Frozen. Applying LightTAR to the
text columns as well changes mean AUC by $-0.11$ points and mean
$R^2$ by $+0.11$ points relative to adapting images alone.

\section{Runtime Measurements}
\label{app:cost}
\paragraph{Representation construction.}
For Table~\ref{tab:cost}, we time tokenization, encoding, fitting the
projection on the training rows, and transforming the training and test
rows. Both methods use \textit{e5-small-v2}, one NVIDIA A40, and eight
CPU threads. We take the median of three runs on the first split of each
of the 20 MulTaBench text datasets, then sum the 20 medians. For the
fine-tuning comparison, we add the measured LoRA adaptation time to the
Frozen encoding time. LoRA adapts the last three layers at rank 16 with
batch size 32 and early stopping within 40 epochs.

\paragraph{Runtime per 1,000 examples.}
For each learner and representation in Figure~\ref{fig:pareto}, we
time fitting and prediction on the first split of each supported
dataset. We take the median of three runs, each with a fresh learner.
Timing excludes learner initialization, loading cached features,
and metric computation. For each dataset, we add the separately measured
representation time and divide by the total number of training and test
rows, then multiply by 1,000. The figure reports the median across datasets.
On MulTaBench image, representation time covers only the image columns,
with five runs per dataset.

The Elo ratings use scores from the five evaluation splits. The
separate timing runs use BF16 and eight ensemble members for TabFM,
a configuration that may differ from the one used for its Elo rating.
\markoff

\markoff

\end{document}